\documentclass[lettersize,journal]{IEEEtran}
\usepackage{amsmath,amsfonts}
\usepackage{bm}
\usepackage{algorithmic}
\usepackage{algorithm}
\usepackage{array}
\usepackage[caption=false,font=footnotesize,labelfont=rm,textfont=rm]{subfig}
\usepackage{textcomp}
\usepackage{stfloats}
\usepackage{url}
\usepackage{verbatim}
\usepackage{graphicx}
\usepackage{multirow}
\usepackage{cite}
\begin{document}

\title{Range-Aided LiDAR-Inertial Multi-Vehicle Mapping in Degenerate Environment}

\author{Zhe Jin, Chaoyang Jiang
        % <-this % stops a space
\thanks{
    This work is supported by the National Natural Science Foundation of China(No.52002026, U20A20333), and the National Key Research and Development Project (No. 2020YFC1512500)(\textit{Corresponding author: ChaoyangJiang}).
    
    The authors are with the School of Mechanical Engineering, Beijing Institute of Technology, Beijing, China, 100081 (email: 3220215054@bit.edu.cn; cjiang@bit.edu.cn;}% <-this % stops a space
%\thanks{Manuscript received April 19, 2021; revised August 16, 2021.}
}



\maketitle

\begin{abstract}
This paper presents a range-aided LiDAR-inertial multi-vehicle mapping system (RaLI-Multi). Firstly, we design a multi-metric weights LiDAR-inertial odometry by fusing observations from an inertial measurement unit (IMU) and a light detection and ranging sensor (LiDAR). The degenerate level and direction are evaluated by analyzing the distribution of normal vectors of feature point clouds and are used to activate the degeneration correction module in which range measurements correct the pose estimation from the degeneration direction.
We then design a multi-vehicle mapping system in which a centralized vehicle receives local maps of each vehicle and range measurements between vehicles to optimize a global pose graph. The global map is broadcast to other vehicles for localization and mapping updates, and the centralized vehicle is dynamically fungible.
Finally, we provide three experiments to verify the effectiveness of the proposed RaLI-Multi. The results show its superiority in degeneration environments.
\end{abstract}


\section{INTRODUCTION}
\IEEEPARstart{M}{ulti-vehicle} simultaneous localization and mapping (SLAM) has been widely used for search and rescue, maintenance investigations, underwater detection, and space exploration \cite{choudhary2017distributed}.
It is a great challenge for a single vehicle to handle the tasks in large-scale and degenerate environments while multi-vehicles working together have great potential to improve mapping accuracy and efficiency. Therefore, multi-vehicle collaborative mapping systems have increasingly attracted attention in recent years \cite{ramachandran2020information}.

Features in large-scale and degenerate environments are usually sparse which leads to great accumulate errors for SLAM systems.
Fortunately, range sensors are invulnerable in degenerate environments in the absence of shading.
On one hand, range constraints are simpler and more efficient than finding loop closures for collaborative mapping; on the other hand, range factors can be easily introduced into a pose graph optimization (PGO) procedure.
Therefore, range-aided multi-vehicle SLAM has great potential to improve the robustness of localization and mapping in degenerate environments.

\subsection{Related works}
The multi-vehicle mapping has two main branches: centralized mapping and decentralized mapping \cite{jamshidpey2021centralization}.
Centralized mapping systems collect and optimize messages from all connected vehicles.
Riazuelo et al. \cite{riazuelo2014c2tam} proposed a typical centralized mapping system in which the expensive map optimization and storage were allocated on a cloud server while a light camera tracking client run on a local computer. Deutsch et al. \cite{deutsch2016framework} further introduced a software framework for real-time multi-vehicle collaborative SLAM which can potentially work with various SLAM algorithms. They both require an external server for the aggregation of data and information feedback, and thus network delays become a hidden problem. In contrast, Dub{\'e} et al. \cite{dube2017online} shifted the master node into one of the vehicles and proposed a fully-integrated online multi-vehicle SLAM system, which saves the long-distance communication but requires a high-performance onboard processor.
Decentralized methods do not rely on a central server and split the computation to each vehicle node. Choudhary et al. \cite{choudhary2017distributed} proposed a set of distributed algorithms for pose graph optimization in which vehicles communicate and exchange relative measurements only when the rendezvous is detected. Different from \cite{choudhary2017distributed}, inter-vehicle communications and pose-graph optimization are real-time implemented in \cite{cieslewski2018data}. Lajoie et al. \cite{lajoie2020door} then extended and improved the above two methods \cite{choudhary2017distributed, cieslewski2018data}, and proposed DOOR-SLAM, a fully distributed SLAM system with an outlier rejection mechanism that can work with less conservative parameters.
The above-mentioned multi-vehicle mapping systems applied loop detection of inter-or-intra vehicles to address data association and have achieved great progress. However, they still cannot work well in degenerate environments, especially when environmental characteristics are similar.

Degeneracy is caused by fewer constraints in some directions, leading to less robustness for state estimation. The characteristics of degenerate environments include lacking geometrical, textural, and/or thermal features.
Zhang et al. \cite{7487211} first proposed a degeneration detection method and separated the degenerate directions in the state space to reduce the influence of the degeneracy in structured environments. Similarly, Hinduja et al. \cite{8968577} only optimized the pose graph in well-constrained directions of the state space. These directions were selected based on a dynamic threshold and real-time updated. Extending the above two methods, Ren et al. \cite{ren2021lidar} proposed a reliable degeneracy indicator that can evaluate the scan-matching performance in off‐road environments. The evaluated degeneracy indicator was then integrated into a factor graph optimization framework. However, these methods \cite{7487211, 8968577, ren2021lidar} only adopted a single sensor and were unable to optimize the degenerate dimension. Khattak et al. \cite{9213865} utilized a visual-inertial odometry and a thermal-inertial odometry to find robust priors for LiDAR pose estimation. One of the two odometry was selected for propagation when LiDAR odometry failed due to degeneration, which can improve the reliability of the pose estimation. Great progress has been achieved in past decades, but robust mapping is still a big challenge in degenerate environments.

Degenerate environments have no influence on the distance observations of range sensors like Bluetooth, ultra-wideband (UWB) ranging sensors, Zigbee and WiFi.
Song et al. \cite{song2019uwb} fused LiDAR and UWB measurements for single-vehicle localization, and allowed the unknown anchors to change their positions. To some extent, it was more robust and resisted degeneration. Similarly, applying more sensors like inertial measurement unit (IMU), light detection and ranging sensors (LiDAR), and camera, Nguyen et al.\cite{nguyen2021viral} performed a comprehensive optimization-based estimator for the state of an unmanned aerial vehicle.
Both methods \cite{song2019uwb, nguyen2021viral} depend on preset anchors which greatly limits their applications for multi-vehicle cases.
Xu et al. \cite{9813359} proposed a decentralized state estimation system, fusing stereo wide-field-of-view cameras and UWB sensors for a multi-vehicle case. Similarly, Nguyen et al. \cite{9655461} proposed a visual-inertial-UWB multi-vehicle localization system that loosely fuses the UWB and visual-inertial odometry data while tightly fusing all onboard sensors. Both methods achieved a great localization improvement but only in small-scale and undegenerate environments.
The current range-aided methods focus on localization with or without anchor beacons but few of them focused on mapping in degenerate environments.

Prior related works on multi-vehicle mapping are rich, but they still have further room for improvement:
1) range-aided multi-vehicle mapping systems with fixed anchors can hardly extend to large-scale environments due to the requirement for numerous anchors while those without anchors still cannot work well in degenerate environments;
2) centralized systems rely on a central server which is vulnerable, and decentralized systems cannot easily achieve a globally consistent map in real time;
3) most anti-degenerate methods ignore the information of degeneration directions or compensate with other sensors like thermal sensors that depend on environmental features;
4) few works cope with the degeneration correction.

\subsection{Contribution}
Considering the above-mentioned problems, we propose the RaLI-Multi: a range-aided LiDAR-inertial multi-vehicle mapping system.
Each vehicle performs a local mapping procedure with IMU integration, LiDAR feature extraction and registration, degeneration detection, and degeneration correction.
Range measurements compensate for the error in the degenerate direction when both the degenerate level and the gap between the LiDAR-inertial odometry and the range measurements exceed their preset thresholds.
The RaLI-Multi dynamically schedules one vehicle as an anchor vehicle which stops and can be viewed as an anchor for range measurement. The anchor vehicle also acts as a temporary central server, which receives local maps, LiDAR-inertial odometry, and range constraints between vehicles to optimize and broadcast the global map that in turn updates the local states of each vehicle.
The main contributions of this paper are as follows:

\begin{itemize}{}{}
\item [1)]
We propose a multi-metric weights LiDAR-inertial front-end, which assigns weights to each feature point and can achieve better odometry in degenerate environments.
\item [2)]
A geometry-based degeneration detection method is proposed as the foundation of the following degeneration correction module, which can online monitor the degeneration level and estimate the corresponding degenerate direction.
\item [3)]
The range-aided degenerate correction module compensates the error of LiDAR-inertial odometry from the degeneration direction which is considered as the main component of the pose estimation error. In this way, we can improve the robustness of the mapping systems in degenerate environments.
\item [4)]
The proposed RaLI-Multi has both advantages of centralized and decentralized methods. All vehicles have communications with the central node and share the same global map. The anchor vehicle plays the role of the central node, which can dynamically shift to other vehicles. Hence, the proposed system is more robust and flexible, which has potential to apply in large-scale degenerate environments.
\end{itemize}

\subsection{Notations and Outline}
We denote a point cloud set captured by a 3D LiDAR sensor on a vehicle as
${\bm{{\cal L}}}$,
and denote a processed feature cloud and normal cloud as ${}^{\cal F}{\cal L}$ and ${}^{\cal N}{\cal L}$, respectively. Range measurements between vehicle $j$ and vehicle $k$ are denoted by
$u_i^{jk} \in {\bm {{\cal U}}}$.
The elements of these sets are presented with subscript of time sequence, e.g., ${\left(  \cdot  \right)_t}$ or ${\left(  \cdot  \right)_i}$.

$\cal X$ is the vehicle state including position, orientation, velocity, etc. For simplicity, we also represent the position of a vehicle as ${\bm x} \in \mathbb{R} {^3}$.
The initial pose transformation between tag vehicles and the anchor vehicle are denoted by
$\bm{{\cal T}} = \left\{ {{{{\cal T}}^1 },{{{\cal T}}^2 },{{{\cal T}}^3 }, \cdots } \right\}$,

\begin{equation}
    {\cal T}^\upsilon = \left[ {\begin{array}{*{20}{c}}
            {{\bm R}^\upsilon}&{{\bm t}^\upsilon}\\
            0&1
    \end{array}} \right] \in SE\left( 3 \right), v = \{1,2,3,\cdots\}
\end{equation}
where ${\bm R}^\upsilon \in SO\left( 3 \right)$ 
and ${\bm t}^\upsilon \in \mathbb{R} {^3}$ 
are the rotation matrix and the translation vector, respectively.
The corresponding quaternion of the rotation is represented by Hamilton notation.

The rest of this paper is organized as follows. Section \ref{sec o} provides the overview. Section \ref{sec m} proposes details of the RaLI-Multi mapping system. Experiment results are shown in Section \ref{sec e}, and conclusions are given in Section \ref{sec c}.

\section{OVERVIEW}
\label{sec o}

\subsection{System Definition}
\label{ss sd}

We propose a range-aid LiDAR-inertial multi-vehicle mapping system, in which all vehicles take the same onboard hardware and software. Each vehicle has an IMU, a LiDAR, a range sensor, a router, and a computing unit. All vehicles have two roles: the anchor role and the tag role, but they cannot be activated simultaneously. If the anchor role is activated, the vehicle acts as an anchor vehicle and vice versa.
During the exploration, one of the vehicles is automatically selected to be the anchor. `The anchor vehicle' also plays the role of the central node of such a multi-vehicle network. All other vehicles are called `the tag vehicle'.

\begin{figure*}[!t]
    \centering
    \includegraphics[width=6.5in]{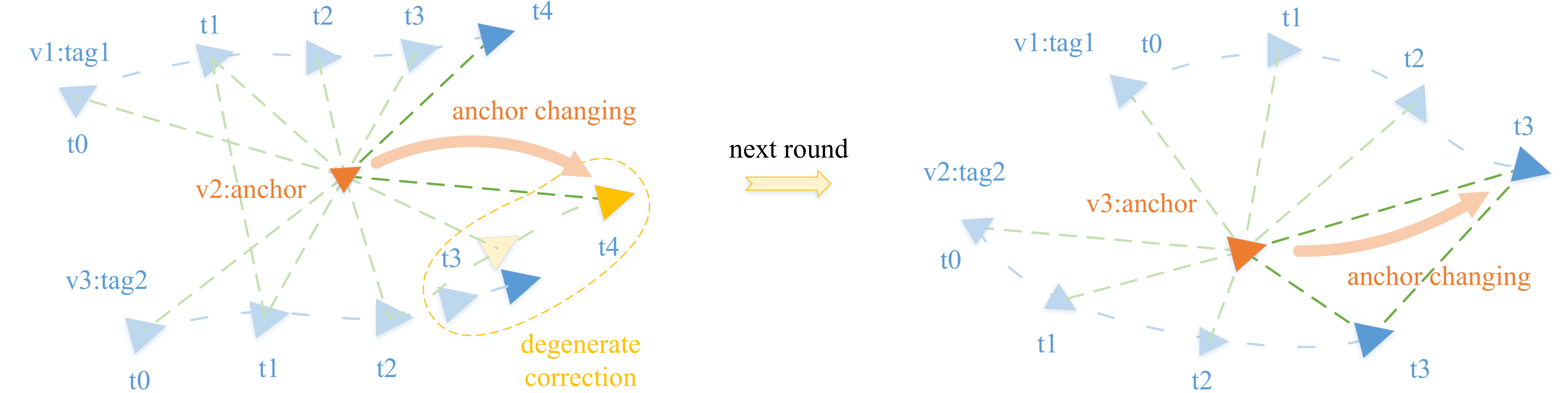}
    \caption{Illustration of two exploration rounds. Blue dotted lines represent the trajectories of tag vehicles and green dashed lines represent range measurements. In the former round, vehicle 2 is selected to be the anchor vehicle and vehicles 1 and 3 are tag vehicles. During exploration, vehicle 3 detects degeneration at the time stamps t3 and t4, which is then corrected by the range measurements between vehicle 2 and vehicle 3. At t4, both tag vehicles finish their exploration. Meanwhile, the anchor role is transferred to vehicle 3, and the latter round starts. t4 in the former round and t0 in the latter is the same time stamp.}
    \label{pf}
\end{figure*}

The RaLI-Multi mapping procedure consists of continuous exploration rounds, as shown in Fig. \ref{pf}. Each round begins with the tag-vehicle exploration and ends with the anchor-vehicle selection. In the first round, a dynamical initialization (see Section \ref{ss di}) is required, which estimates the relative transformation between the global frame (the coordinate frame of the initial anchor vehicle) and local frames (coordinate frames of the initial tag vehicles).
When all tag vehicles finish their exploration, the role of the anchor and the central node shifts from one vehicle to another. A tag vehicle finishes its exploration in the current round if one of the following three events is triggered: 1) the Received Signal Strength Indicator (RSSI) of the communication is less than a pre-defined threshold; 2) the distance with the anchor vehicle exceeds a pre-defined value; 3) the environment around the tag vehicle has been fully explored.

\subsection{Problem Formulation}
\label{ss pf}

We aim to reconstruct 3-D maps for large-scale environments with degeneration via multiple vehicles.
Our main ideas are applying range observations between the anchor vehicle and all tag vehicles for degeneration correction, and utilizing communications and range observations for the improvements of the global mapping and pose estimation of all vehicles.
Consequently, this work mainly focuses on the following three problems:
\begin{itemize}{}{}
\item [1)]
How to correct the localization and mapping for degenerate cases?
\item [2)]
How to globally optimize the mapping and the pose estimation of all vehicles in such a RaLI-Multi mapping system?
\item [3)]
How to dynamically select the role of the anchor vehicle?
\end{itemize}

\subsection{System Overview}
\label{ss so}

\begin{figure}
    \centering
    \includegraphics[width=3.5in]{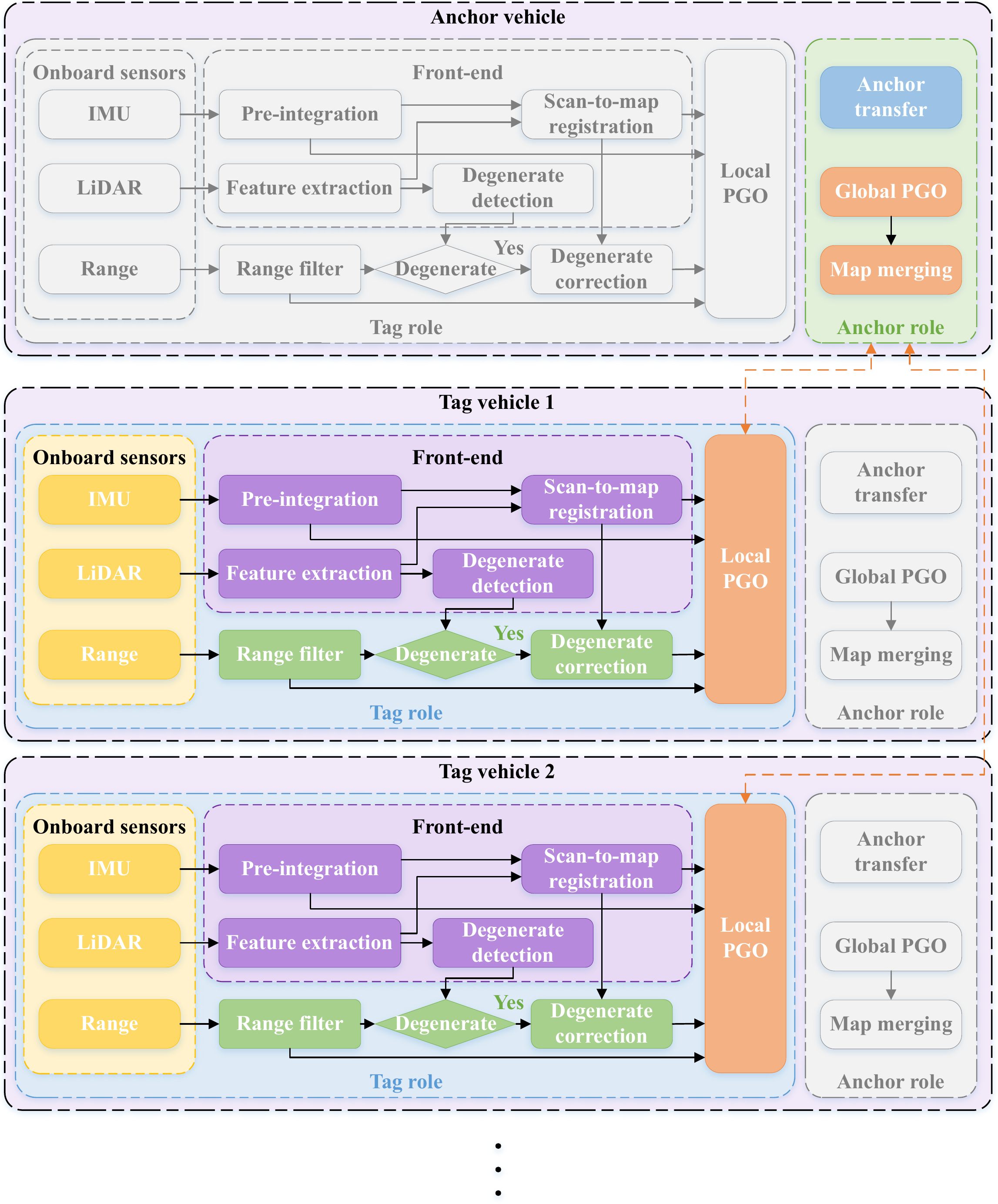}
    \caption{System structure of the tag-vehicle exploration. At the end of each round, a new vehicle is selected to be the anchor vehicle via the anchor transfer module on the current anchor vehicle. Tag roles are then triggered for the rest tag vehicles.}
    \label{wf}
\end{figure}

The structure of the tag-vehicle exploration is shown in Fig. \ref{wf}.
The anchor vehicle stays stationary while all other vehicles, i.e., the tag vehicles, explore the environment. Each tag vehicle performs a LiDAR-inertial odometry, a degeneration detection module, a degeneration correction module with the range measurements from the anchor vehicle, and a local PGO. With the information received from tag vehicles, the anchor vehicle optimizes the poses of tag vehicles and the global map. When all tag vehicles finish their exploration, one of the vehicles is dynamically selected to be the anchor vehicle in the anchor transfer module, followed by the next exploration round.

Each tag vehicle first preprocesses the raw data received from its onboard IMU, LiDAR, and the range sensor. The observations of IMU are pre-integrated (see Section \ref{ss ipi}). Features are extracted from the point cloud of LiDAR (see Section \ref{ss fe}), and the range measurements are pre-smoothed. Then, the LiDAR-inertial front-end takes pre-integrated IMU states as an initial guess to perform scan-to-map registration (see Section \ref{ss s2m}). Meanwhile, the features are used for degeneration detection (see Section \ref{gbdd}), and the range constraints are used for degeneration correction (see Section \ref{ss dc}). Finally, the corrected LiDAR odometry, IMU pre-integration, and range constraints are jointly optimized via a local PGO in the back end. With the above procedure, the first question mentioned in Section \ref{ss pf} is answered.

During a round, local data from each tag vehicle are sent to the anchor vehicle after local PGO for global PGO. If tag vehicles have stable range signals between each other and the RSSI is stronger than the pre-set threshold, these range measurements are also added to global PGO (see Section \ref{ss urr}). The anchor vehicle then performs an incremental global optimization and map merging (see Section \ref{igpmm}).
After global optimization, the anchor vehicle broadcasts the global map and the optimized states of all tag vehicles to each tag vehicle. In this way, we solve the second problem mentioned in Section \ref{ss pf}.

Like the classical frontier-based exploration method \cite{613851}, we define frontiers as the boundary between known free space and unknown space. If no frontiers exist, the environment is regarded as fully explored. When all tag vehicles finish their exploration, the current anchor vehicle starts to select the next anchor vehicle (see Section \ref{ss dars}). The frontiers of each tag vehicle are combined if they are close to each other. The vehicle that is closest to the center of the largest frontier is selected as the new anchor. In such a manner, the third problem mentioned in Section \ref{ss pf} is figured out.

\begin{figure}
    \centering
    \subfloat[initial constraint]{\includegraphics[width=1.1in]{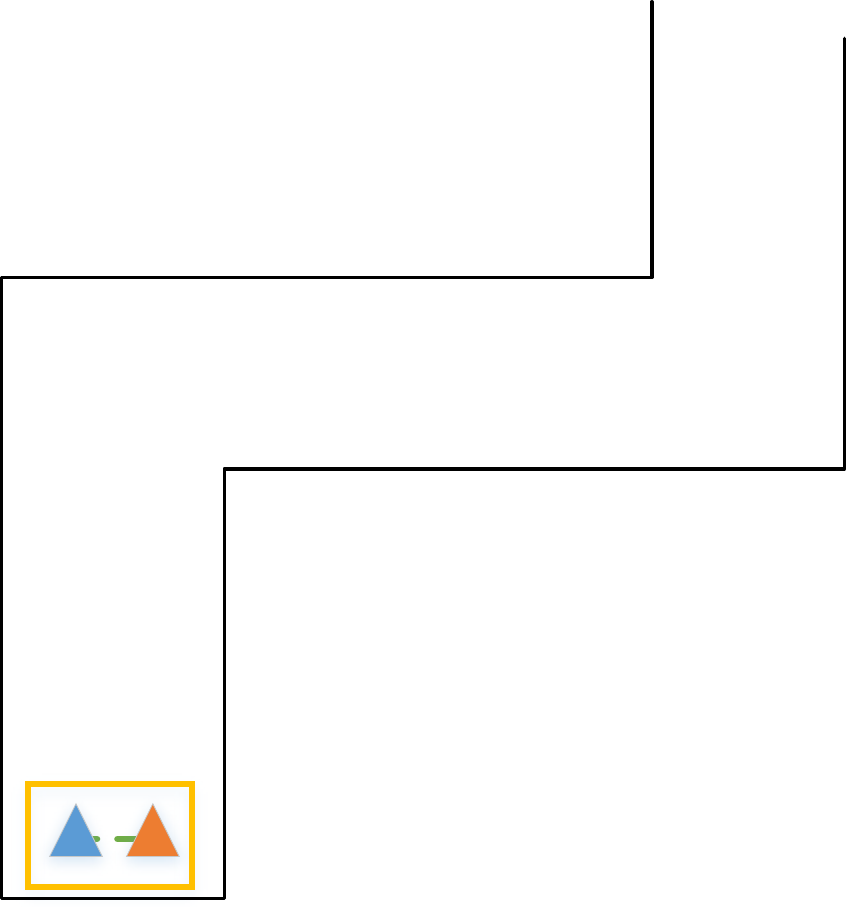}}\hspace{2pt}
    \subfloat[round 1]{\includegraphics[width=1.1in]{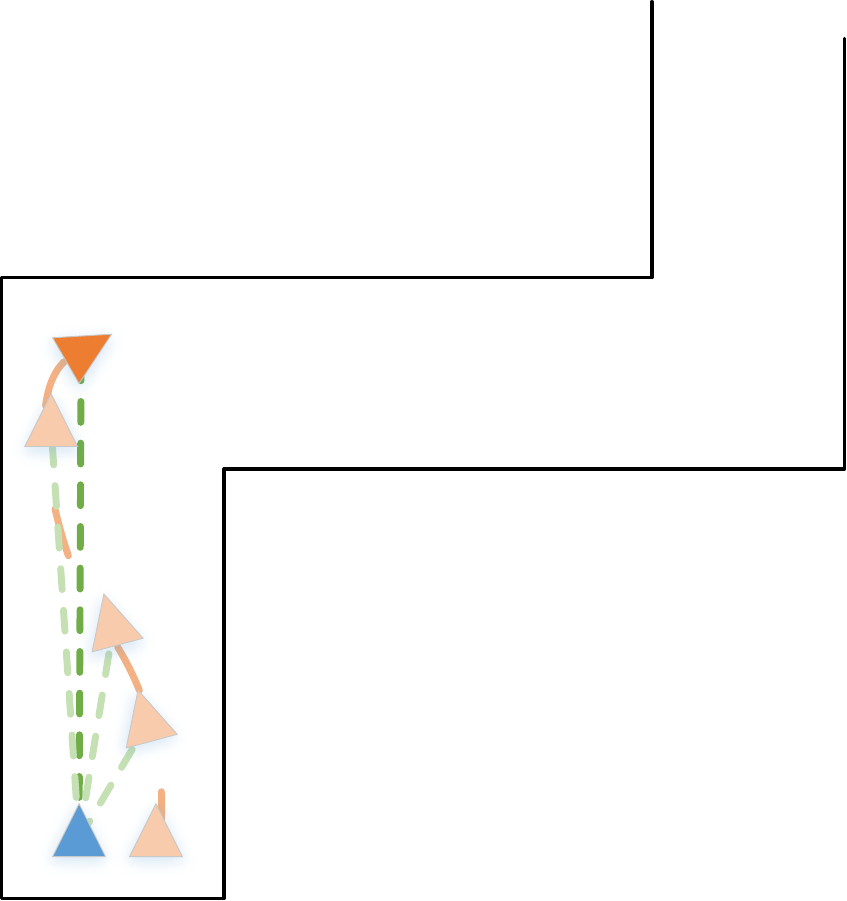}}\hspace{2pt}
    \subfloat[round 2]{\includegraphics[width=1.1in]{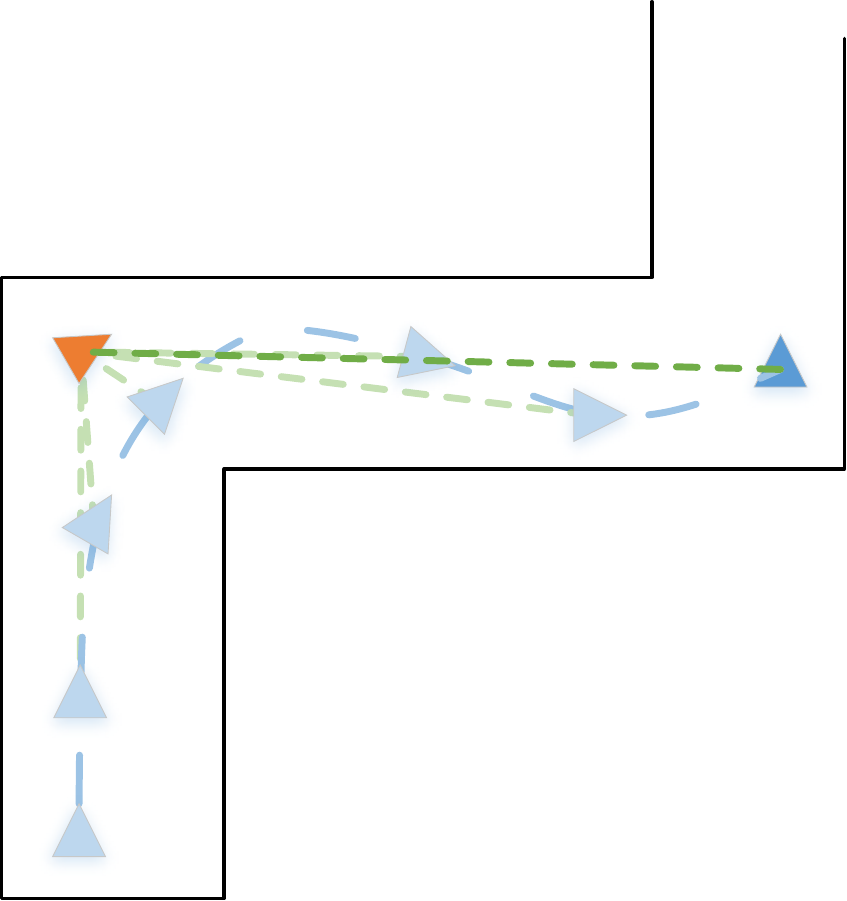}}
    \caption{Illustration of a two-vehicle mapping system exploring a corridor-like environment. The orange and blue triangles represent two vehicles. The yellow rectangle is the initial relative pose constraint. Green dashed lines are range measurements. Orange and blue dotted lines are the trajectories of two vehicles.}
    \label{f 5x3}
\end{figure}

To easily understand the workflow of the RaLI-Multi, a two-vehicle example is shown in Fig. \ref{f 5x3}, which tells the procedure of how the two vehicles explore a corridor-like environment.
The global coordinate frame is fixed on the local coordinate frame of the blue triangle, i.e., the initial anchor vehicle. Before mapping, an initial relative pose prior between two vehicles consisting of a range measurement and pre-set parameters is added to the pose graph, as shown in the yellow rectangle. Next, the tag vehicle, i.e., the orange triangle,  begins to explore around, as shown in Fig. \ref{f 5x3} (b). During this period, range measurements between two vehicles constrain the poses of the tag vehicle and reduce the influence of degeneration. Meanwhile, the anchor vehicle receives poses and corresponding LiDAR point clouds of the tag vehicle to perform initialization and incremental global pose graph optimization. After the tag vehicle finishes its exploration, the anchor vehicle transfers optimized results back to the tag vehicle. Finally, two vehicles exchange the roles of tag and anchor to start the next round of exploration, as shown in Fig. \ref{f 5x3} (c).

\section{RaLI-Multi MAPPING SYSTEM}
\label{sec m}

\subsection{LiDAR-inertial Odometry}
\label{sec lio}

\subsubsection{IMU Pre-integration}
\label{ss ipi}
IMU pre-integration was first introduced by Forster et al. in \cite{forster2015imu} to reduce recomputation when changing linearization points.
However, it can also be seamlessly integrated whether visual-inertial, LiDAR-inertial, or other inertial-related pipelines under the holistic framework of factor graphs. Here, we use the same procedure as \cite{forster2015imu}, and ignore the details of IMU pre-integration.

\subsubsection{Feature Extraction}
\label{ss fe}
As pointed out in Ye et al. \cite{8793511}, edge points can hardly improve the results of the LiDAR-inertial odometry in practice.
Additionally, extracting edge points is time-consuming, and we find that edge points bring larger errors than planar points due to the less horizontal resolution of LiDAR sensors.
As a result, we only extract planar points.

\begin{figure}
    \centering
    \subfloat[]{\includegraphics[width=1.5in]{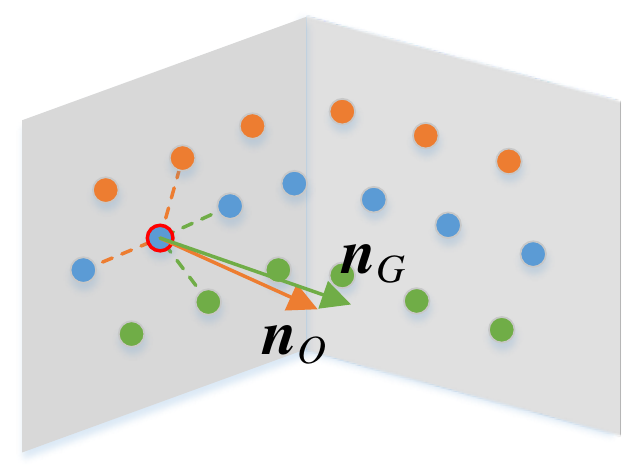}}\hspace{12pt}
    \subfloat[]{\includegraphics[width=1.5in]{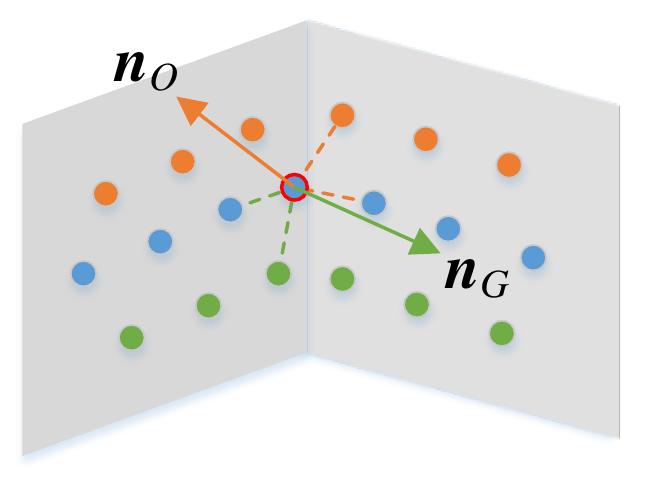}}
    \caption{Illustration of the feature extraction. Points on the same ring are represented as the same color. For simplicity, only three different rings are shown. The point with a red border represents the candidate point. (a) The candidate point is selected to be a planar point whose neighbor region is in a plane. (b) The candidate's point is rejected.}
    \label{f fe}
\end{figure}

We first downsample the raw point cloud and call the four nearest points of each candidate point as the neighbor points, which are found by the $k$-d tree, shown in Fig. \ref{f fe}. The distances between each neighbor point and the candidate point should be less than double of the downsample resolution. Furthermore, the neighbor points should distribute in three different rings. The candidate point and two neighbor points are on the same ring, as shown in the blue points in Fig. \ref{f fe}. The rest two neighbor points are in the nearest rings, as shown in the orange and green points respectively in Fig. \ref{f fe}. Two unit normal vectors, ${\bm n}_G$ and ${\bm n}_O$, as shown in the green and orange arrows, are the cross products of two vectors, i.e., dash lines with corresponding colors in Fig. \ref{f fe}. Finally, the angle between the two normal vectors is calculated via their dot product:
$\theta  = {\cos ^{ - 1}}\left\langle {{{\bm n}_G},{{\bm n}_O}} \right\rangle $.
The point is selected as a planar point if $\theta$ is less than a pre-set threshold. Otherwise, this point will be rejected.
The normal vector of the planar point can be defined as the unit vector of the summation of two normal vectors, i.e.,
$\bm n_i = \frac{{{\bm n_G} + {\bm n_O}}}{{\left| {{\bm n_G} + {\bm n_O}} \right|}}$.

\subsubsection{Scan-to-Map Matching with Multi-Metric Weights}
\label{ss s2m}
We propose a group of multi-metric weights of LiDAR points and apply the relative transformation obtained from IMU pre-integration as the initial guesses to update the front end. The source cloud is the planar feature cloud ${}^{\cal F}{\cal L}$ extracted in the former part and the target cloud is the submap consisting of the nearest $N_{kf}$ keyframes in the local map of each vehicle. Our scan matching module then estimates the pose of the current point cloud in the submap coordinate system.

For each iteration, we first transform a point to the submap frame. The neighbor points in the submap are determined by the nearest neighbor search within a pre-set range threshold with the origin as the center of the current point. Then, we estimate the normal vector ${\bm n}_j$ the same as extracting planar feature points.
The optimal pose ${\cal X}_i$ is given by the resolution of the point-to-plane distance cost function,

\begin{equation}
\begin{array}{l}
    {r_{\cal L}}\left( {{{\cal X}_i},{{\bm{{\cal L}}}_i}} \right) = 
    \mathop {{\rm{argmin}}}\limits_{{{\cal X}_i}} \mathop \sum \limits_j \rho \left( {{\omega _j}\left\langle {{\bm{R}_i}{\bm{p}_j} + {\bm{t}_i} - \bm{p}_j^{center},{\bm{n}_j}} \right\rangle } \right)
\end{array}
\end{equation}
where
$\rho \left(  \cdot  \right)$ 
is a Huber lost function and ${\bm{R}_{i}}$ and ${\bm{t}_{i}}$ are the rotation matrix and the translation vector of ${\cal X}_i$, respectively. ${\bm p}_j$ is the current point and ${\bm p}_j^{center}$ is the mass centroid of the neighbor points. The multi-metric weight ${\omega}_j$ is

\begin{equation}
    {\omega _j} = {\eta _r}\omega _j^{range} + {\eta _n}\omega _j^{neighbor} + {\eta _k}\omega _j^{kinematic},
\end{equation}
where
\begin{equation}
    \omega _j^{range} = \frac{1}{{1 + {e^{ - \frac{{2.5}}{{{l_{Q3}}}}\left( {{r_j} - {l_{Q2}}} \right)}}}},
    \label{eq wr}
\end{equation}

\begin{equation}
    \omega _j^{neighbor} = \left\{ {\begin{array}{*{20}{c}}
            {\frac{{n_j^{neighbor}}}{{{N_{neighbor}}}},}&{{n_j^{neighbor}} < {N_{neighbor}}}\\
            {1,}&{{n_j^{neighbor}} \ge {N_{neighbor}}}
    \end{array}} \right.,
    \label{eq wn}
\end{equation}

\begin{equation}
    \omega _j^{kinematic} = \left\{ {\begin{array}{*{20}{c}}
            {{{\cos }^{ - 1}}\left\langle {{p_j},{n_j}} \right\rangle  \cdot {r_j},}&{\delta {\theta _j} > {\theta _{th}}}\\
            {0,}&{else}
    \end{array}} \right.,
    \label{eq wk}
\end{equation}
$\eta_r$, $\eta_n$ and $\eta_k$ are normalized weights (taken 0.5, 0.2 and 0.3 for all experiments, respectively). $\omega_j^{range}$ enhances the influence of far points. In (\ref{eq wr}), $r_j$ represents the range of the current point ${\bm p}_j$. $l_{Q2}$ and $l_{Q3}$ are the second and the third quartile of all ranges in current feature points. $e$ is a constant. $\omega_j^{neighbor}$ guarantees that the current point locates in a sphere area with a high point density. In (\ref{eq wn}), $n_j^{neighbor}$ is the number of the neighbor points found by the nearest neighbor search and the search radius is usually set to be double of the point cloud downsample resolution. $N_{neighbor}$ is a pre-set threshold relating to the sample resolution. $\omega_j^{kinematic}$ is designed for large rotation conditions. In (\ref{eq wk}), $\delta \theta_j$ represents the rotation angle of the IMU pre-integration result and can be defined from quaternion
$\delta {\bm{q}_j} = \left( {w,x,y,z} \right)$
as
$\delta {\theta _j} = {\tan ^{ - 1}}\left( {\sqrt {{x^2} + {y^2} + {z^2}} ,w} \right)$. $\theta _{th}$ is a pre-defined rotation angle threshold.

\subsubsection{Keyframe Selection}
We find in experiments that common keyframe selection methods \cite{9681177, 9341176} including both distance-based and rotation-based methods are unstable in an indoor or narrow environment, especially at the corner of a corridor. The distance-based keyframe selection methods are hard to obtain a keyframe at the corner of a corridor leading to less robustness when passing through the corner. The rotation-based methods can easily induce distortion of point clouds due to vehicle vibration.

To efficiently select keyframes in indoor or narrow environments, we consider the overlap of two point clouds through Octree \cite{meagher1982geometric}, which is faster than $k$-d tree in voxel searching. After transferring the current scan to the frame of the last keyframe, if the distance between a point in the current scan and its closest point in the last keyframe is less than double of the downsampled resolution, the point is labeled as overlap. If the ratio of overlap points in the current scan is less than a pre-set threshold, we select the current scan as a keyframe.

\subsection{Geometry-based Degeneration Detection}
\label{gbdd}

\begin{figure}
    \centering
    \subfloat[]{\includegraphics[width=1.7in]{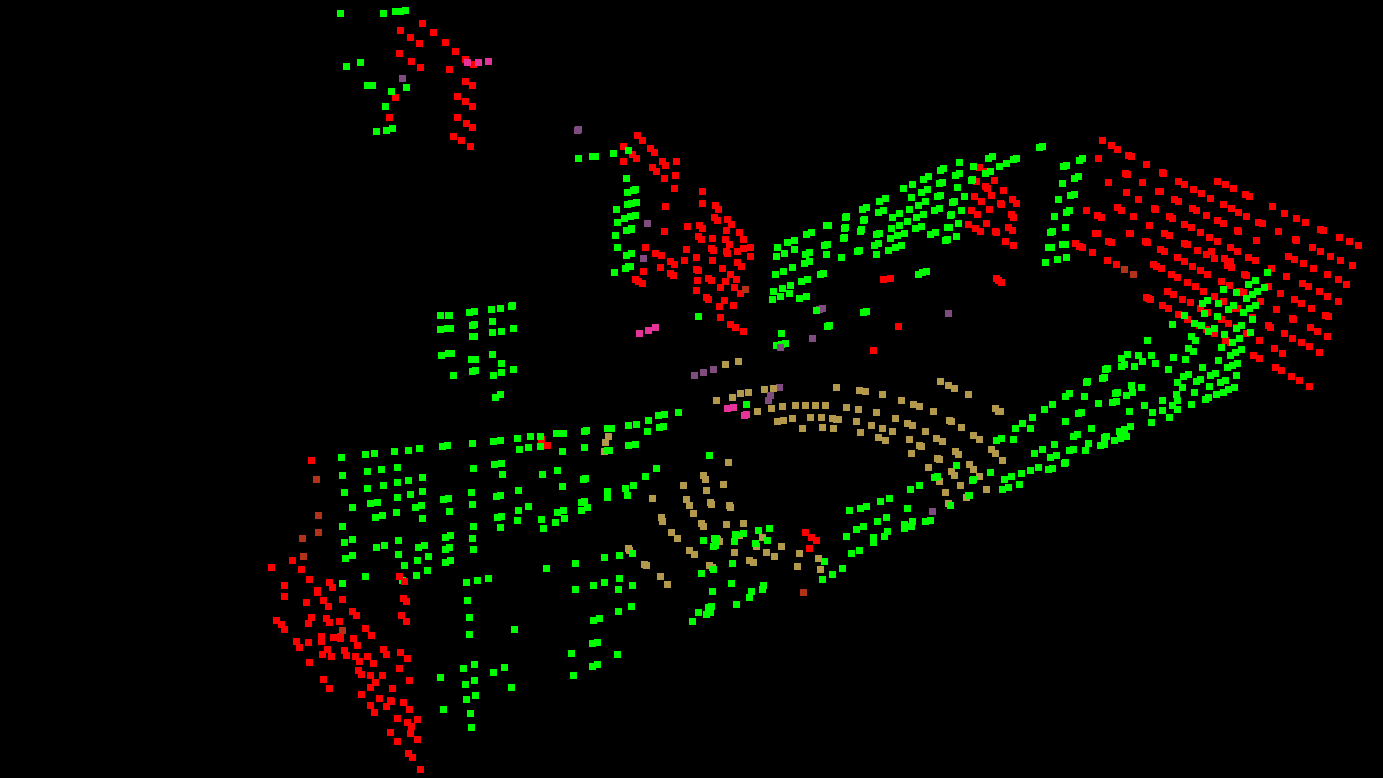}}\hspace{2pt}
    \subfloat[]{\includegraphics[width=1.7in]{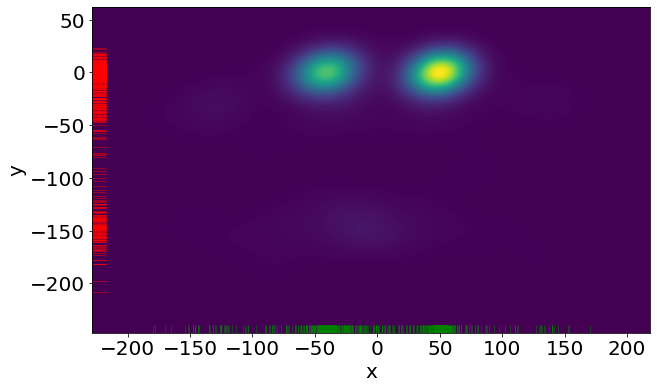}}\quad
    \subfloat[]{\includegraphics[width=1.7in]{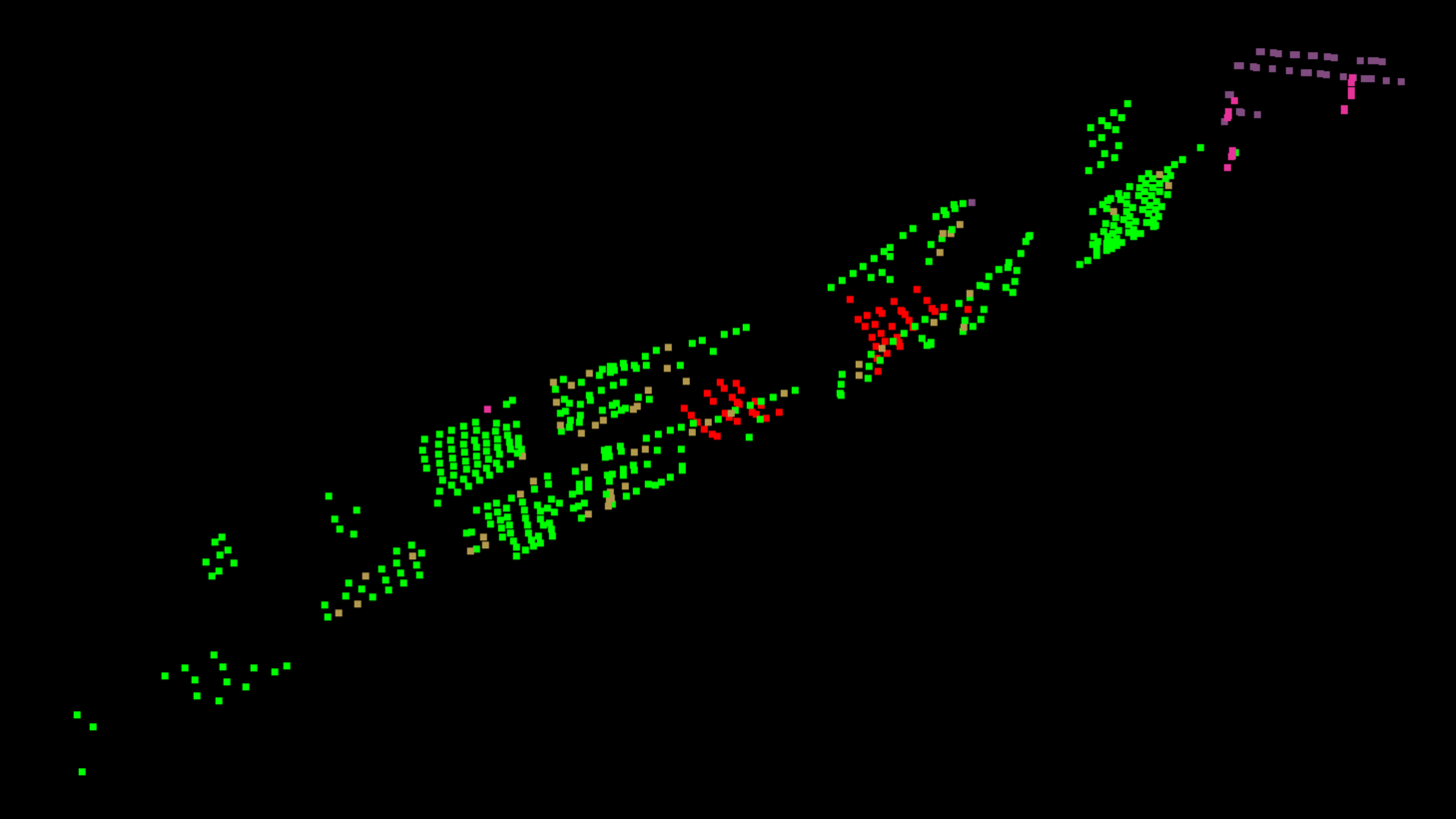}}\hspace{2pt}
    \subfloat[]{\includegraphics[width=1.7in]{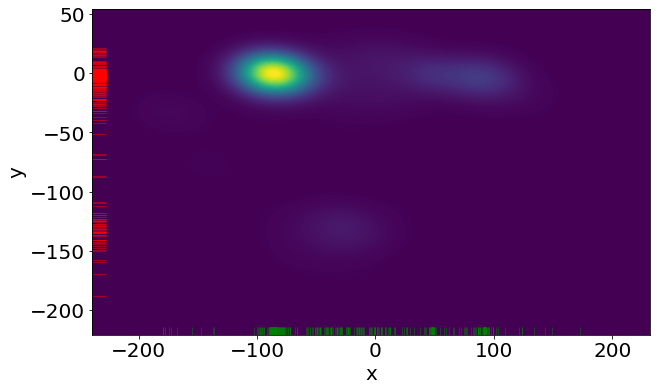}}
\caption{Point clouds with random colors in (a) and (c), represent different clusters of normal vectors. (b) and (d) are corresponding normal vectors projected onto a two-dimensional plane. Stick marks in red and green colors on the coordinate axes represent the distribution of raw data on different axes, respectively. The brighter the color, the more normal vectors there are in this area.}
\label{f gbdd}
\end{figure}

Firstly, we take two examples to present the degeneration detection method: a non-degenerate environment and a degenerate environment, as shown in Fig. \ref{f gbdd} (a) and (c). The colors of points represent different clusters of normal vectors and are generated randomly.
In Fig. \ref{f gbdd} (a), red and green points represent mutually perpendicular walls, while brown points are the ground plane, and other colors, such as pink and purple for example, can be treated as noise points. However, in Fig. \ref{f gbdd} (c), green points represent the wall that occupies most of the view and red points are the ground plane. Purple points are the other wall at an angle of approximately 45 degrees to the green-point wall. In order to better visualize the degeneracy of the environment, we project normal vectors from the three-dimensional sphere coordinate system to a two-dimensional plane coordinate system by applying the Mercator-like projection method. The results are shown in Fig. \ref{f gbdd} (b) and (d), respectively. From the density maps, it is simple to identify walls perpendicular to the floor from yellow and green areas in both scenarios. However, ground points in both pictures and purple-points wall in Fig. \ref{f gbdd} (c), colored in light blue, are not obvious due to fewer number, which are located around (20, -140) in Fig. \ref{f gbdd} (b) and (80, 0), (-30, -130) in Fig. \ref{f gbdd} (d).

According to the above examples, we find that normal vectors in a degenerate environment are highly characterized. These vectors can be classified into finite clusters and the number of them in different clusters varies widely. Then, we formulate the degeneracy by analyzing the distribution of normal vectors through the Principal Components Analysis module (PCA). We treat normal vectors set as the normal cloud ${}^{\cal N}{\cal L}$, and the covariance matrix ${\bf \Sigma}_n$ of ${}^{\cal N}{\cal L}$ is calculated as follows,

\begin{equation}
{{\bf{\Sigma }}_n} = \frac{1}{N_{{}^{\cal N}{\cal L}}}\mathop \sum \limits_{i = 1}^{N_{{}^{\cal N}{\cal L}}} \left( {{{\bm{n}}_i} - \bar {\bm{n}}} \right){\left( {{{\bm{n}}_i} - \bar {\bm{n}}} \right)^ \top }
\end{equation}
where ${\bar {\bm{n}}}$ is the mass center of ${}^{\cal N}{\cal L}$ and $N_{{}^{\cal N}{\cal L}}$ is the number of points in ${}^{\cal N}{\cal L}$. Then, eigenvalues
${\lambda _1} \ge {\lambda _2} \ge {\lambda _3} \ge 0$
are determined by eigenvalue decomposition of ${\bf \Sigma}_n$.

The degeneration can occur in all directions separately or simultaneously. To simplify the problem, we make the following two assumptions: 1) Due to our vehicles moving on the ground, we assume that the LiDAR sensors will always observe the ground plane and vehicles will not degenerate in the vertical direction; 2) Unlike exploring the open terrain such as grassland, a desert, a lake, etc. where there are no sufficient constraints in all horizontal directions for the LiDAR odometry, we assume that only one direction is mainly degenerated in the horizontal plane. The typical examples include corridor, tunnel, underground passage, and so on.
Thus, we merely consider the smallest two eigenvalues, i.e., $\lambda _2$ and $\lambda _3$, and the distribution of the normal cloud ${}^{\cal N}{\cal L}$ can be determined by the \textit{degenerate degree} $\sigma_{deg}$ inspired by \cite{hackel2016fast}: $\sigma_{deg} = \frac{{{\lambda _2}}}{{{\lambda _3}}}$, and the \textit{degenerate direction} is the eigenvector of the smallest eigenvalue, i.e., $\bm e_3$. If the \textit{degenerate degree} $\sigma_{deg}$ is less than a threshold, the environment is considered as degeneration.

\subsection{Range Constraints for Degenerate Correction}

\subsubsection{Range Residuals}
\label{ss urr}

Diverse sensors can be used for range measurement, such as UWB, Zigbee, WiFi, light sensors, and so on. All measurements are noisy. Considering the measurement noise, we online smooth the raw data of the range observations for a past time horizon with a least square smoother. Then, the residuals of range measurements between vehicle $j$ and vehicle $k$ at timestamp $i$ can be formulated as,

\begin{equation}
    {r_u}\left( {{{\cal X}_{i}^{{v_j}}},{{\cal X}_{i}^{{v_k}}},{u_i^{jk}}} \right) = \left| {{\bm x}_i^{{v_j}} - {\bm x}_i^{{v_k}}} \right| - {u_i^{jk}} + {\eta _{{u_i^{jk}}}}
\end{equation}
where ${\cal X}_{i}^{v_j}$ and ${\cal X}_{i}^{v_k}$ represent states of two vehicles at the timestamp $i$ obtained from the LiDAR-inertial odometry and $u_i^{jk}$ represents the corresponding smoothed range measurement.
${\eta _{{u_i^{jk}}}} \sim {\cal N}\left( {0,\sigma _{{u_i^{jk}}}^2} \right)$
represents the noise following a zero-mean Gaussian noise.

\subsubsection{Degenerate Component Correction}
\label{ss dc}

\begin{figure}
    \centering
    \includegraphics[width=3.5in]{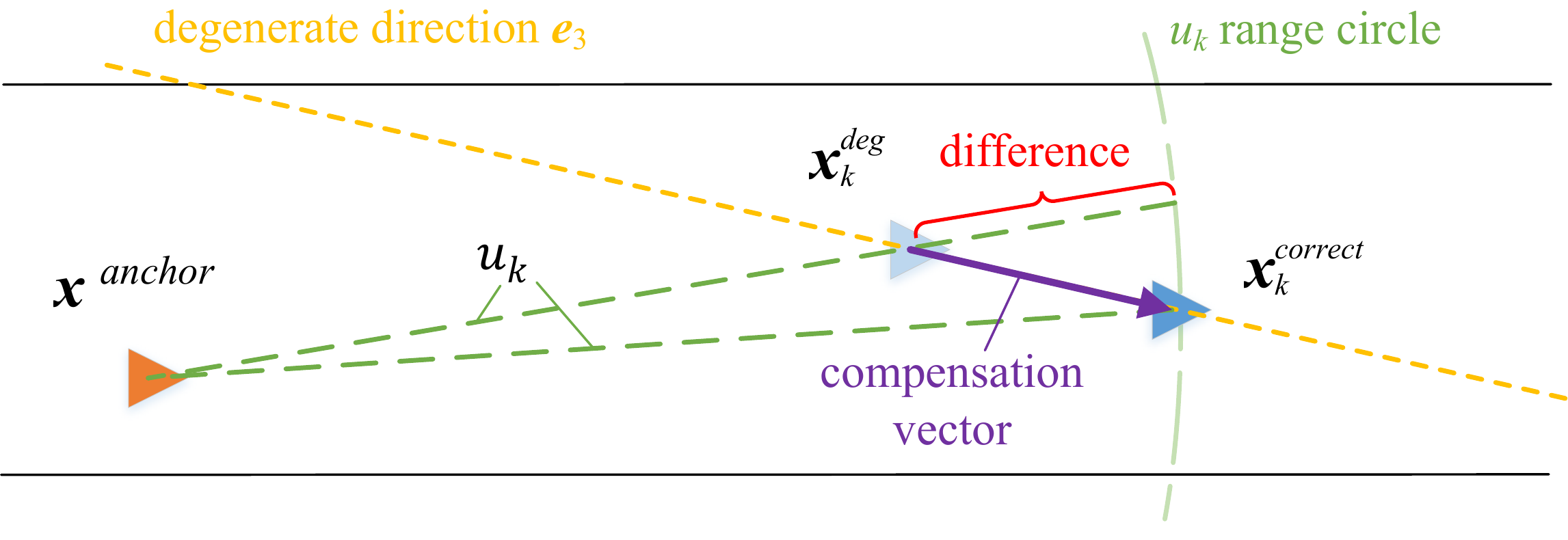}
    \caption{Degeneration correction through the range information. The orange triangle is a static anchor vehicle and the rest two triangles represent a common tag vehicle where the light blue triangle is the degenerate state and the deep blue triangle is the corrected state. The green dotted line is the range circle and the green dashed lines are the corresponding radius. The red curly bracket shows the difference between the LiDAR-inertial odometry and the range measurement. The yellow dashed line and the purple arrow represent the degenerate direction and the compensation vector at timestamp $k$, respectively.}
    \label{f dc}
\end{figure}

According to the distribution of features proposed in Section \ref{gbdd}, the environmental degeneration can be real-time monitored. If the degeneration is detected, and the gap between the LiDAR-inertial odometry and the range measurement exceeds a threshold, we can apply the range observation to reduce the position drift, based on the \textit{degenerate direction} calculated in Section \ref{gbdd}. As shown in Fig. \ref{f dc}, the corrected state ${\bm x}_k^{correct}$ should be located on the circle centered on the anchor vehicle with radius ${u_k}$, which represents the range measurement between the anchor vehicle and the tag vehicle at the timestamp $k$. We omit the superscript of ${u_k}$ for simplicity. We view the state estimation ${\bm{x}_k^{deg}}$ as a vector with ${\bm{x}^{anchor}}$ as the origin. Since the gap is mainly due to the degeneration, the error vector of the estimated state ${\bm{x}_k^{deg}}$ is considered on the \textit{degenerate direction}, which is represented by the unit eigenvector $\bm e_3$. We denote the magnitude of the error by $s$.
Then, constraining the problem on the XY coordinate, we can obtain the error $s{\bm e_3}$, which we call the compensation vector, from the equation

\begin{equation}
    \left| {\bm x_k^{deg} - \bm x_k^{anchor} + s{\bm e_3}} \right| = {u_k} + {\eta _{{u_k}}}. 
\end{equation}
With $s{\bm e_3}$, we can correct the influence of the degenation and obtain
\begin{equation}
\bm{x}_k^{correct}=\bm{x}_k^{deg}+s{\bm e_3}.
\end{equation}

\subsection{Dynamical Initialization}
\label{ss di}
Before globally optimizing the pose graph, an anchor vehicle needs to unify the coordinate systems of all vehicles. To reduce the computational burden in the following exploration, we estimate the transformations between the global frame and each local frame in the first round of exploration and fix these transformations in the following rounds.

At the beginning of the first exploration round, an initial anchor vehicle is randomly selected from all vehicles. The global frame is defined as the local frame of this initial anchor vehicle. As the RaLI-Multi is dynamically centralized, the anchor vehicle may be different in various exploration missions and therefore is the global frame.

During initialization, each tag vehicle performs the odometry as described in Section \ref{sec lio}. Meanwhile, the anchor vehicle receives the odometry and local maps published by each tag vehicle, range measurements between two vehicles, and pre-set initial pose priors. When the size of local maps exceeds a pre-set threshold, the anchor vehicle starts to perform the initialization as follows,

\begin{equation}
\begin{array}{l}
    \mathop {{\rm{argmin}}}\limits_{{}^L\bm{{\cal X}},\bm{{\cal T}}} \left\{ {\mathop {\sum\limits_{v \in {\rm{ }}V} {\left\| {r_{\cal L}^{v}\left( {^L{\bm {{\cal X}}^v}}, \bm{{\cal L}}^v \right)} \right\|_{{\bf{P}}_{\cal L}^{ - 1}}^2} } + r_{scan2map}^v \left( \bm{{\cal T}} \right)} \right.\\
    \left. { + \sum\limits_{u_i^{jk} \in \bm{{\cal U}},v_{j,k} \in {\bm{{\cal V}}_a}} {\left\| {{r_u}\left( {{}^L{{{\cal X}}_{i}^{{v_j}}},{}^L{{{\cal X}}_{i}^{{v_k}}},{{{\cal T}}^{{v_j}}},{{{\cal T}}^{{v_k}}},u_i^{jk}} \right)} \right\|_{{\bf{P}}_u^{ - 1}}^2} } \right\}
\end{array}
\end{equation}
where
$ \bm{{\cal V}} $
represents the set of tag vehicles and
$ \bm{{\cal V}}_a $
represents the set of vehicles in the RaLI-Multi, including an anchor vehicle and all tag vehicles.
${r_{\cal L}^{v}\left( {^L{\bm {{\cal X}}^v}}, \bm{{\cal L}}^v \right)}$
is the LiDAR-inertial odometry residual and
${}^L \bm {{\cal X}}^v$
is a set of local poses of vehicle $v$.
$r_{scan2map}^v \left( \bm{{\cal T}} \right)$
is the scan-to-map registration residual of all tag vehicles where the scan represents the point cloud captured by an anchor vehicle and the map is a local map from the corresponding tag vehicle.
${}^L{{{\cal X}}_i^{{v_j}}},{}^L{{{\cal X}}_i^{{v_k}}}$
are local poses of two vehicles with range constraints $u_i^{jk}$.
${r_u}\left( {{}^L{{{\cal X}}_{i}^{{v_j}}},{}^L{{{\cal X}}_{i}^{{v_k}}},{{{\cal T}}^{{v_j}}},{{{\cal T}}^{{v_k}}},u_i^{jk}} \right)$
is the range constraint between two vehicles and is defined as,

\begin{equation}
\begin{split}
    {r_u}&\left( {{}^L{{{\cal X}}_i^{{v_j}}},{}^L{{{\cal X}}_i^{{v_k}}},{{{\cal T}}^{{v_j}}},{{{\cal T}}^{{v_k}}},u_i^{jk}} \right) = \\
    &\left| {{{\bm R}^{{v_j}}}{}^L{{{\bm x}}_i^{{v_j}}} + {\bm t}^{v_j} - {{\bm R}^{{v_k}}}{}^L{{{\bm x}}_i^{{v_k}}}} - {\bm t}^{v_k} \right| - u_i^{jk} + {\eta _{{u_i^{jk}}}}.
\end{split}
\end{equation}

\subsection{Incremental Global PGO and Map Merging}
\label{igpmm}
During the exploration of tag vehicles, an anchor vehicle serves as a temporal base station, processing incremental global PGO and map merging. Messages transferring from tag vehicles to an anchor vehicle include vehicle poses optimized by local PGO, corresponding LiDAR point clouds, and range measurements between two vehicles, tag-to-tag or tag-to-anchor. The optimization progress is similar to scan-to-map matching described in Section \ref{sec lio} and range constraints in Section \ref{ss urr}. To reduce the computational burden of an anchor vehicle, we reduce the iteration number of the global optimization if no degeneration occurs, and only optimize poses in the current exploration round when there are no loop closures at the system level. At the end of each exploration round, an anchor vehicle publishes the global map and optimized poses to corresponding tag vehicles.
Hence, all vehicles share the same global map.

\subsection{Dynamically Anchor Role Selection}
\label{ss dars}

After all tag vehicles finish their exploration, the next anchor vehicle is selected. Finish conditions are listed as three cases described in Section \ref{ss sd}. In the third case, if all frontiers in the exploration area of a tag vehicle have been examined, the exploration of this tag vehicle in the current round is finished. Then, the selection of next anchor vehicle is determined by frontiers. The current anchor vehicle combines frontiers received from each tag vehicle and finds the largest frontier area. Finally, the vehicle closest to the center of the largest frontier is selected as the new anchor.

\section{EXPERIMENTS}
\label{sec e}

\subsection{Implementation}

\begin{figure}
    \centering
    \subfloat[exp1]{\includegraphics[width=1in]{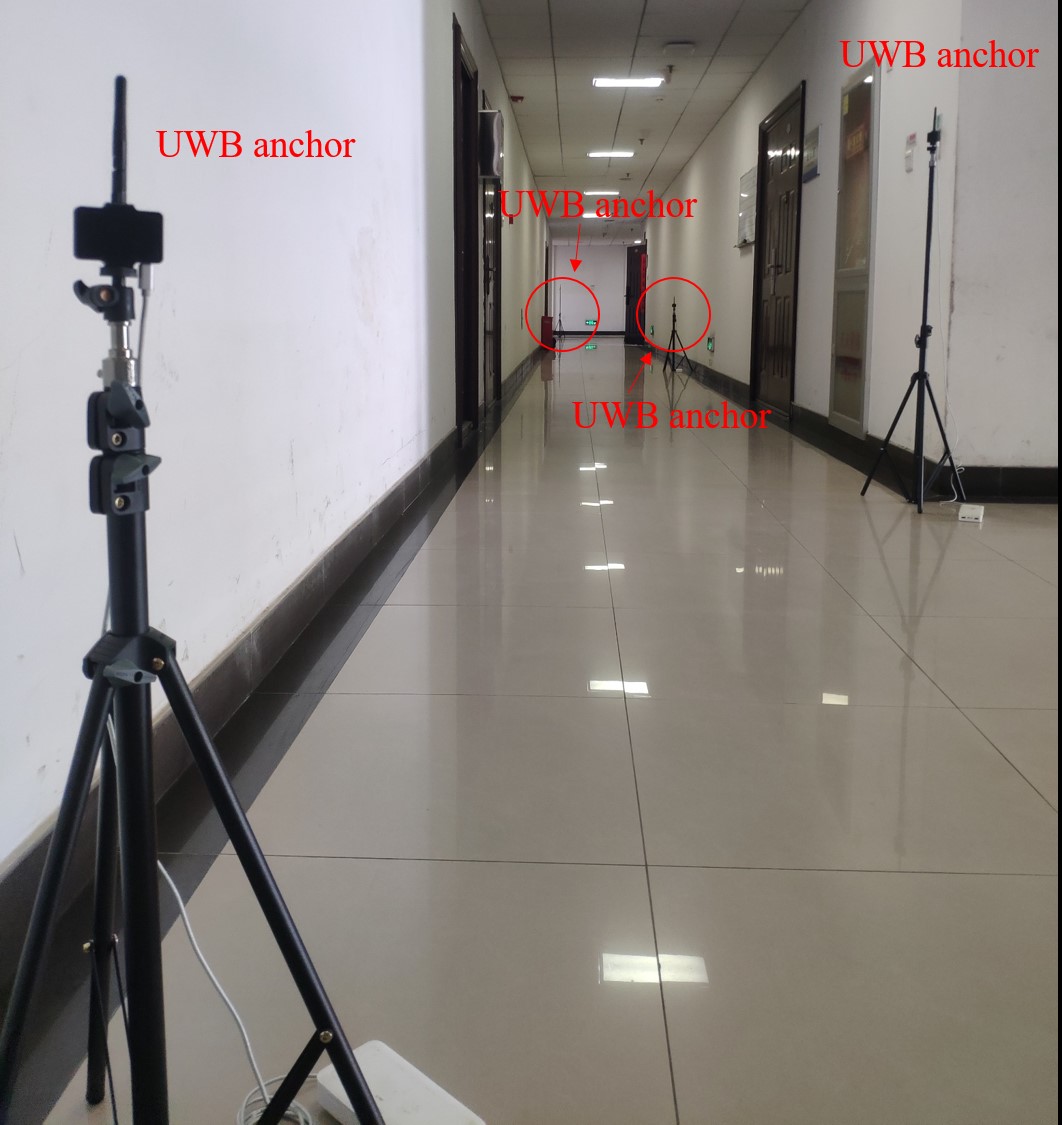}}\hspace{2pt}
    \subfloat[exp2]{\includegraphics[width=1in]{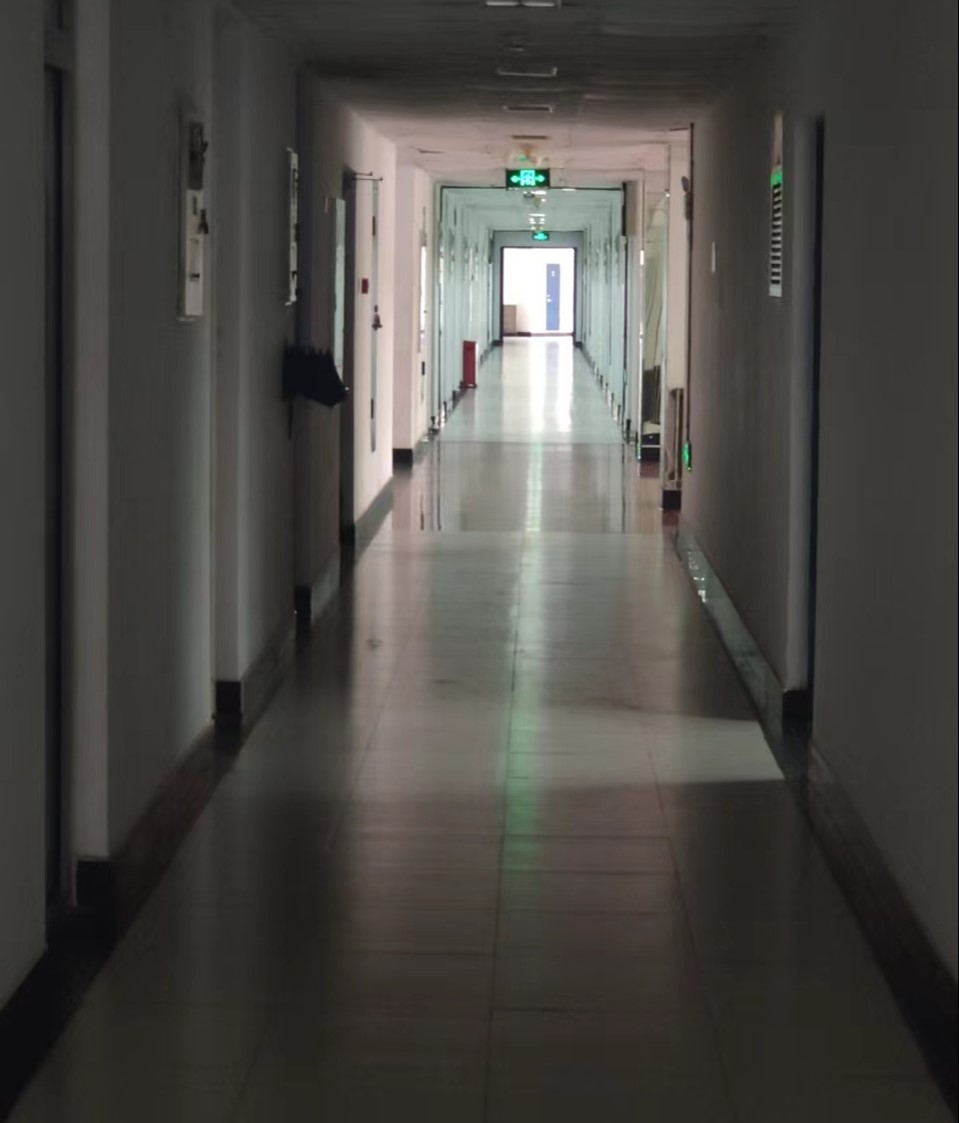}}\hspace{2pt}
    \subfloat[exp3]{\includegraphics[width=1in]{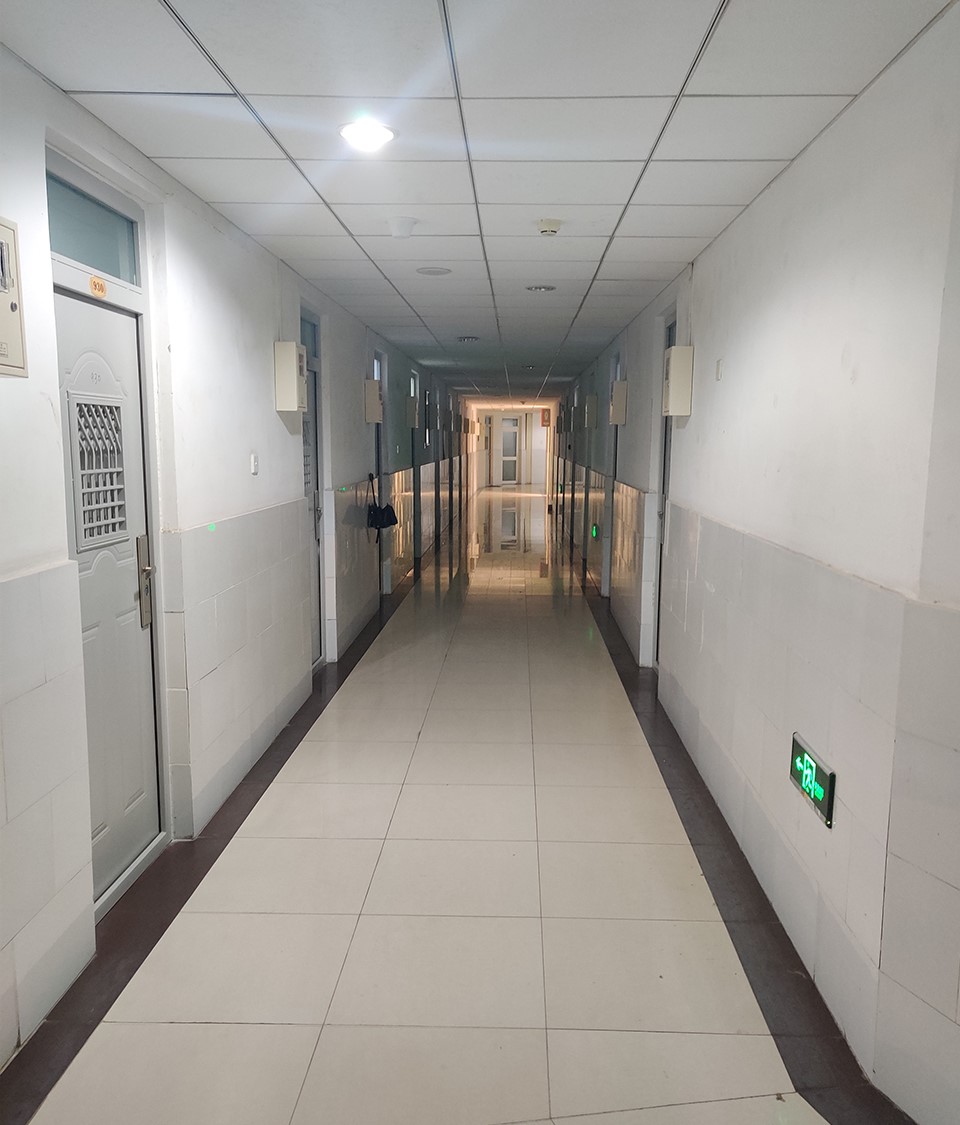}}\quad
    \subfloat[hardware setup]{\includegraphics[width=3.1in]{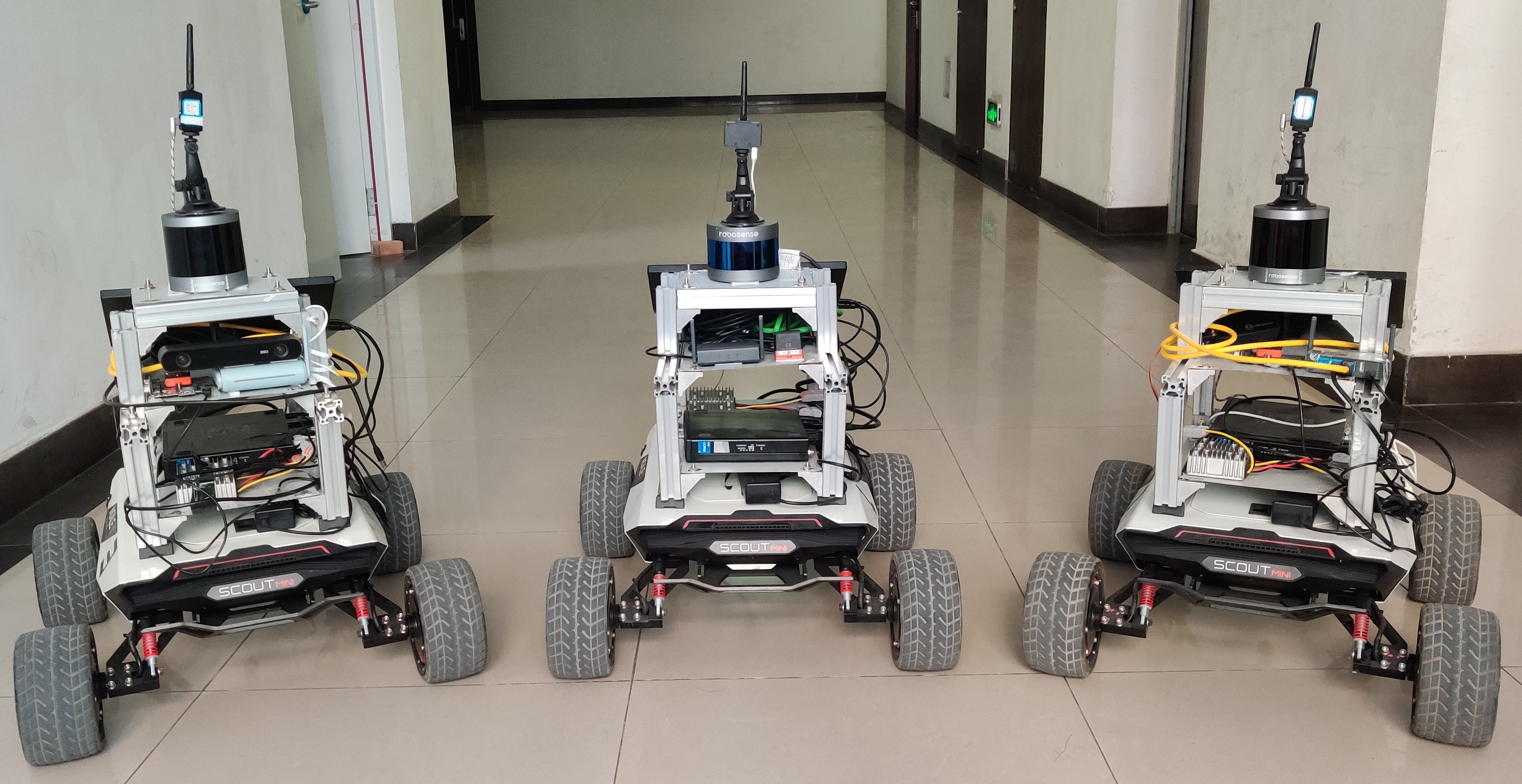}}
    \caption{(a) The first experimental scenario with UWB anchors. (b) and (c) are experimental scenarios in exp2 and exp3, respectively. (d) Hardware setup of the RaLI-Multi.}
    \label{f uwba}
\end{figure}

We perform three experiments to evaluate the proposed methods: the LiDAR-inertial odometry analysis (exp1), the RaLI-Multi with two vehicles in a long corridor-like environment (exp2), and the RaLI-Multi with three vehicles in a complex environment (exp3). The first experiment is mainly designed for evaluating the accuracy of the LiDAR-inertial odometry. UWB anchors applied in exp1 are shown in Fig. \ref{f uwba} (a) to provide a reference trajectory. Fig. \ref{f uwba} (b-c) show the scenarios in exp2 and exp3. Fig. \ref{f uwba} (d) shows unmanned ground vehicles with LiDAR (RoboSense RS-LiDAR-16), UWB (Nooploop LinkTrack P-B), and IMU (Xsens). In exp2 and exp3, the UWB node on each vehicle is applied for inter-vehicle distance measurement, and we use the UWB model proposed by Nguyen et al. \cite{nguyen2018robust}. Specifically, our experimental vehicles equip with differential steering and spring-damped suspension and there is high friction between rubber tires and tiled floors. These reasons lead to vehicles being prone to sharp changes in height when steering, which probably results in large errors in the Z-axis.

We implemented the proposed RaLI-Multi in C++ and Robots Operating System (ROS). We use the GTSAM\cite{2012Factor} framework for the local and global PGO. The Levenberg-Marquardt algorithm is used to solve the pose graph optimization. Trajectory errors in exp1 are calculated by EVO\cite{grupp2017evo} and point cloud map errors are estimated by point-to-mesh distance in CloudCompare\footnote{https://github.com/CloudCompare/CloudCompare} after a coarse-to-fine alignment.

\subsection{Degeneration Analysis}

\begin{figure}[!t]
\centering
    \subfloat[degenerate degree in exp1]{\includegraphics[width=3in]{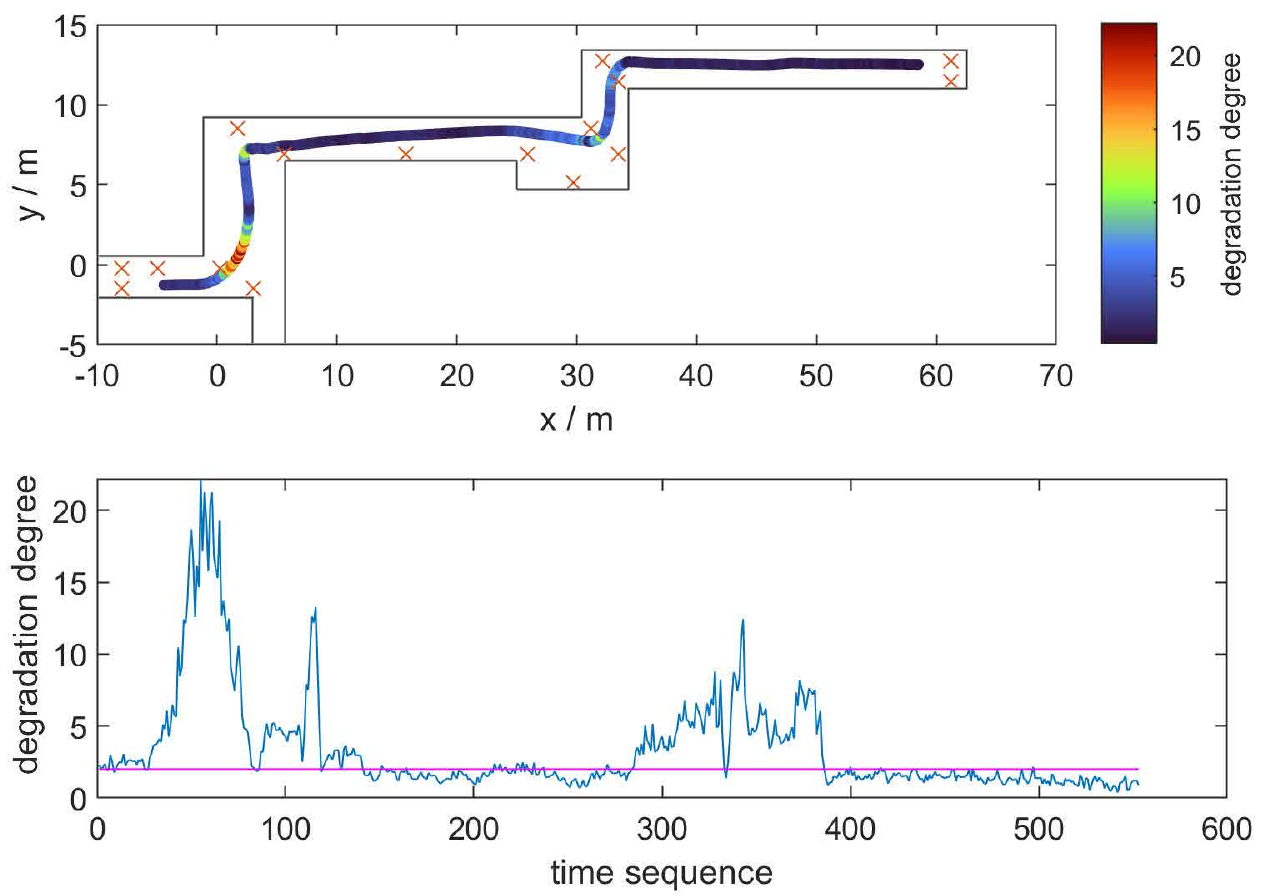}}\quad
    \subfloat[degenerate degree in exp2]{\includegraphics[width=3in]{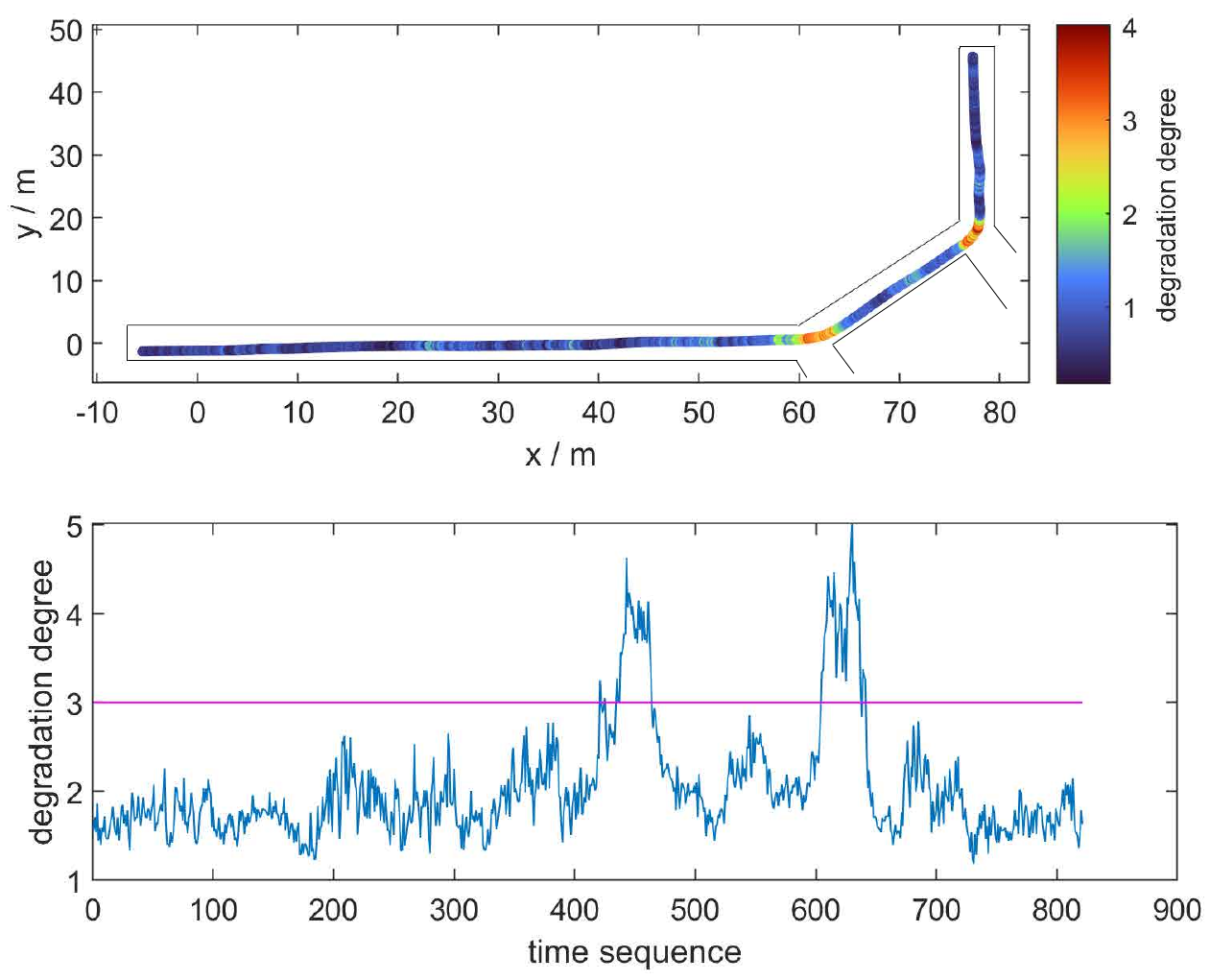}}
\caption{Degeneration degree in exp 1 and 2.}
\label{f deg}
\end{figure}

We first analyze the degenerate level in exp1 and exp2 as shown in Fig. \ref{f deg} (a) and Fig. \ref{f deg} (b), respectively. The lower the degenerate degree, the higher the degenerate level. Red crosses in Fig. \ref{f deg} (a) are locations of UWB anchors which ensure that each vehicle receives at least four range measurements anywhere along the route. In order to clearly illustrate the degenerate level, we present both the spatial and temporal dimensions. In the spatial part, degenerate values are higher in a corner than in a straight corridor, and the longer the corridor, the lower the degenerate value. As shown in the x-y coordinate system, degenerate values in corners are colored in red or green while straight corridors are mostly in blue. In the temporal part, we find that degenerate values of exp1 are higher than that of exp2. It corresponds with that exp1 is less degenerate than exp2. In our experiments, we define a place that is degenerate when its degenerate value is smaller than 3.0.

\subsection{LiDAR-inertial Odometry Evaluation}

\begin{table}[!t]
\begin{center}
\caption{Trajectory APE of different methods in exp1.}
\label{0traj}
\begin{tabular}{ c c c c c }
\hline
& DLO & A-LOAM & FAST-LIO2 & ours\\
\hline
mean & 0.437628 & 0.670686 & 1.516598 & $\textbf{0.213296}$\\
\hline
median & 0.162350 & 0.298089 & 0.152092 & $\textbf{0.115924}$\\
\hline
RMSE & 0.615530 & 0.964350 & 2.941966 & $\textbf{0.308127}$\\
\hline
\end{tabular}
\end{center}
\end{table}

\begin{figure}[!t]
\centering
\subfloat[trajectory results in XY plane ]{\includegraphics[width=3in]{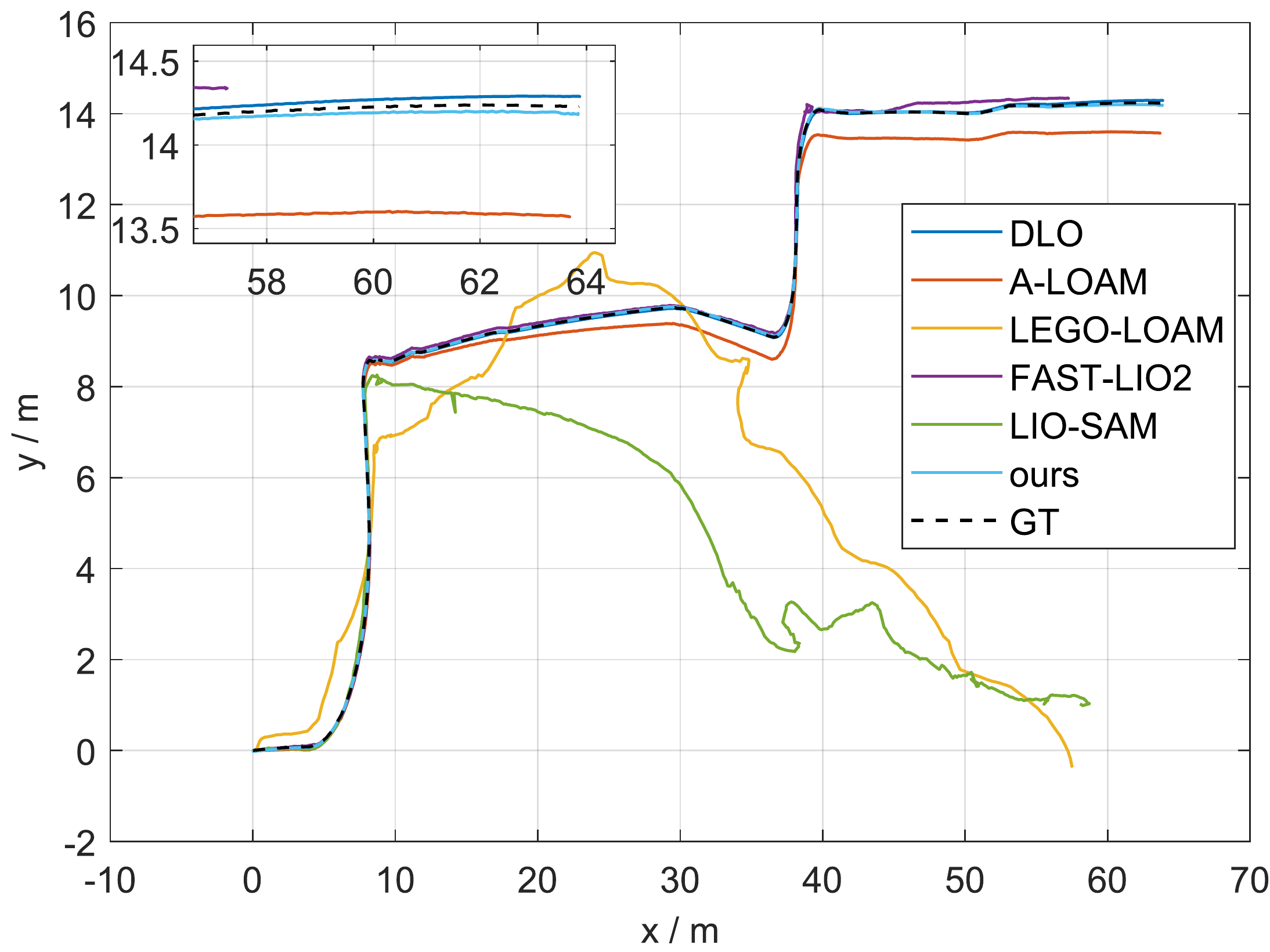}}\quad%1.8
\subfloat[z-axis errors and orientation results ]{\includegraphics[width=3.4in]{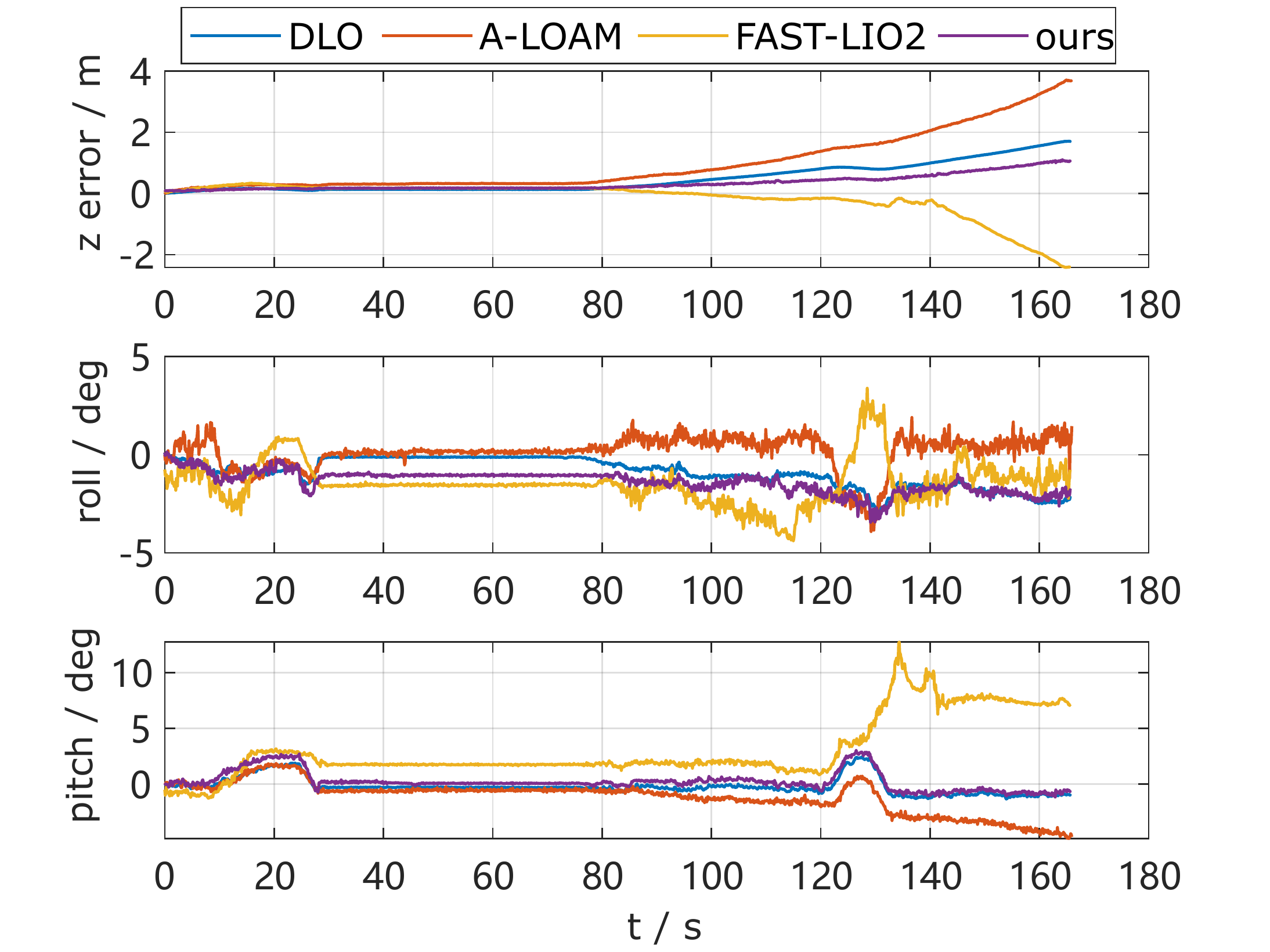}}%\vspace{12pt}
\caption{Trajectory and orientation results of different methods in exp1.}
\label{0xy}
\end{figure}

Firstly, we discuss the performance of the LiDAR-inertial odometry in exp1. As a result of the lack of ground truth, we place UWB anchors around the environment to measure the distances between a tag and anchors via Time of Flight (TOF). We then calculate the tag coordinates with these distances.
Before the experiment, we pre-deploy UWB beacons as shown in Fig. \ref{f uwba} (a). These UWB beacons are placed at different heights to monitor the height variation of these vehicles. Theoretically, it is enough for three UWB beacons to estimate the position of a target. However, considering the robustness and preciseness of the proposed distributed localization system, we redundantly arrange beacons and make sure that each vehicle can receive more than three UWB beacon signals wherever in exp1.

\begin{figure}[!t]
\centering
\subfloat[FAST-LIO2]{\includegraphics[width=3.2in]{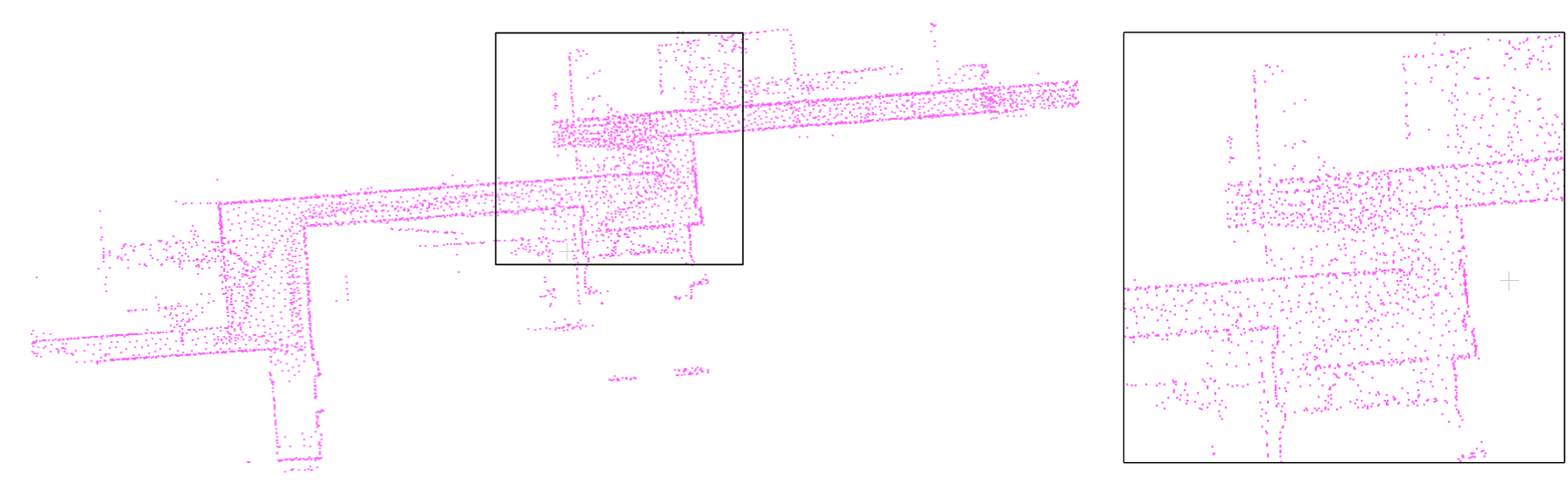}}\quad%\hspace{10pt}1.7-1.5
\subfloat[ours]{\includegraphics[width=3.2in]{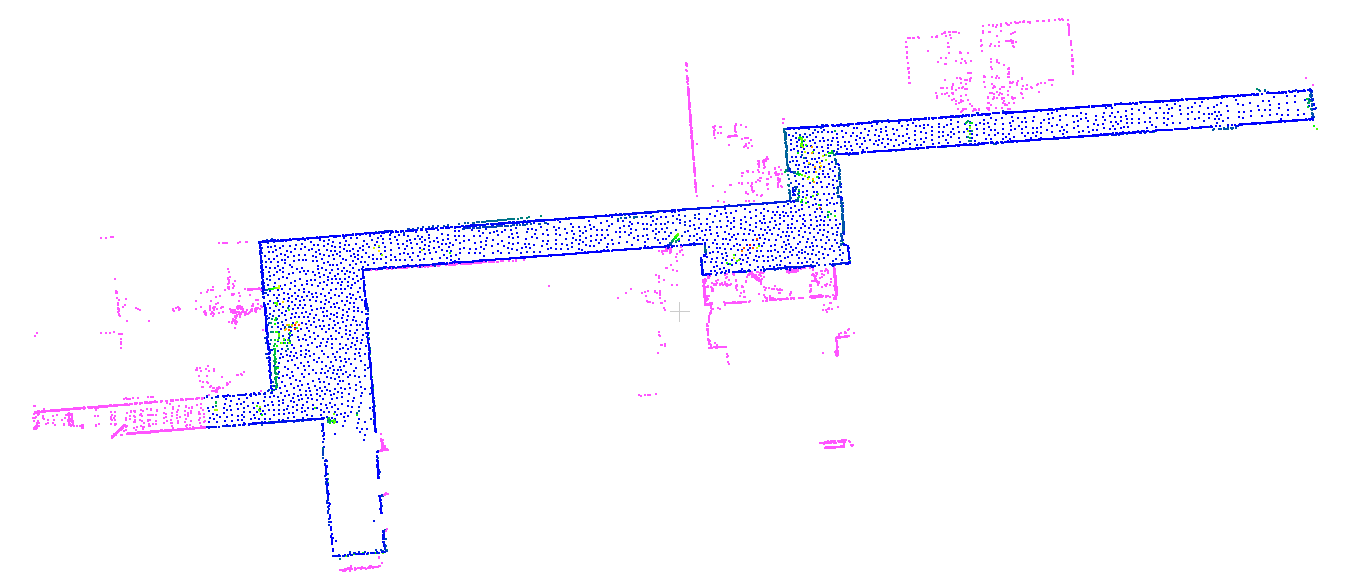}}
\caption{Mapping results of FAST-LIO2 and the proposed method. The drifted area is detailed in partial enlargement.}
\label{0mfast}
\end{figure}

We compare the proposed LiDAR-inertial odometry with DLO \cite{9681177}, A-LOAM\footnote{https://github.com/HKUST-Aerial-Robotics/A-LOAM}, LeGO-LOAM \cite{8594299}, FAST-LIO2 \cite{9697912} and LIO-SAM \cite{9341176}, as shown in Fig. \ref{0xy} (a) and (b). We can see that LeGO-LOAM and LIO-SAM failed in exp1 where the former degenerates at the beginning of this experiment and the latter degenerates when entering the corridor located near (10, 8.5) in Fig. \ref{0xy} (a). As a result, we exclude two of them in Fig. \ref{0xy} (b). Among the rest four methods, FAST-LIO2 and A-LOAM also drifts in various degrees. Similar to LIO-SAM, A-LOAM starts to drift next to (10, 8.5) in Fig. \ref{0xy} (a), especially on the y-axis. In contrast, FAST-LIO2 drifts after the last corner, in the vicinity of (50, 14) in Fig. \ref{0xy} (a). The odometry degeneration mostly occurs on the x-axis and the drifted point cloud map can be seen in Fig. \ref{0mfast} (a). Moreover, both A-LOAM and FAST-LIO2 have an obvious deviation in the z-axis and pitch. Comparing DLO and the proposed method, both of them resist the degeneration and show little difference to the reference trajectory, labeled ground truth (GT) in Fig. \ref{0xy} (a). However, DLO drifts more on the z-axis than the proposed method. Trajectory errors compared with reference are illustrated in table \ref{0traj} through EVO APE (Absolute Pose Error),
${APE_i} = \left\| {{E_i}} \right\|$,
where
${E_i} = {x_{est,i}} - {x_{ref}}$.
MEAN and RMSE in table \ref{0traj} can be calculated from the APEs of all timestamps as follows,

\begin{equation}
{\rm{MEAN}} = \frac{1}{N}\sum\limits_{i = 1}^N {APE_i}
\end{equation}

\begin{equation}
{\rm{RMSE}} = \sqrt {\frac{1}{N}\sum\limits_{i = 1}^N {APE_i^2} }
\end{equation}

\begin{figure}[!t]
\centering
\subfloat[DLO]{\includegraphics[width=2.1in]{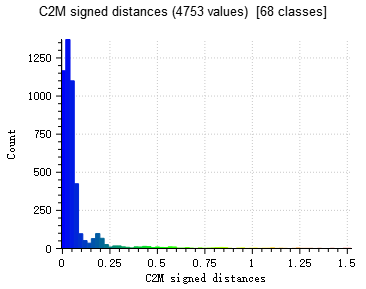}}\quad
\subfloat[A-LOAM]{\includegraphics[width=2.1in]{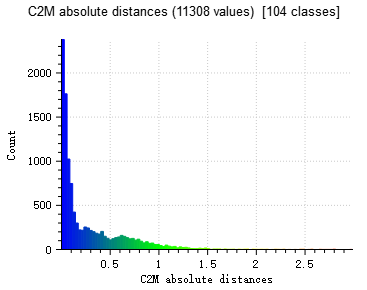}}\quad
\subfloat[ours]{\includegraphics[width=2.1in]{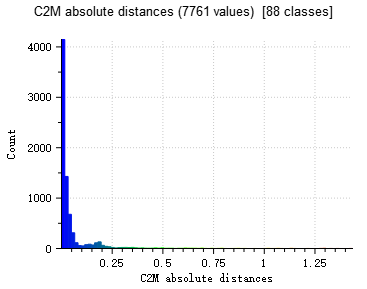}}\quad
\caption{C2M distance results in exp1.}
\label{0map}
\end{figure}

We evaluate the mapping results of DLO, A-LOAM, FAST-LIO2, and the proposed method. The ground truth of the point cloud map is established through CATIA 3D modeling and geometric dimensions are measured by laser measuring instruments. A point cloud map of the proposed method, shown in Fig. \ref{0mfast} (b), is labeled in different colors where magenta points are outlier points excluded in Cloud-to-Mesh (C2M) distance estimation and the rest colors corresponding to C2M distances, the same color as in Fig. \ref{0map} (c). Considering large drifts in the mapping results of FAST-LIO2, we ignore further evaluation. Results of the rest three methods are shown in Fig. \ref{0map} and table \ref{tmap}. On one hand, as a result that A-LOAM records every LiDAR scan into the map, there are more pedestrian points in the final map than DLO and the proposed method. These outlier points are difficult to remove. Thus, the C2M distances statistic results of A-LOAM will be higher than the other two methods though we downsample its point cloud. On the other hand, A-LOAM suffers from degeneration as analyzed in the former paragraph. Both reasons lead to C2M distances of A-LOAM being much higher than those of DLO and the proposed method. Comparing Fig. \ref{0map} (a) and (c), the distribution of C2M distances of the proposed method is closer to zero than DLO. Moreover, the bar closest to zero occupies over half of the whole points while DLO is less than one-quarter.

\subsection{Evaluation in a Long Corridor like Environment}

\begin{figure}[!t]
    \centering
    \subfloat[trajectory results in XY plane]{\includegraphics[width=3in]{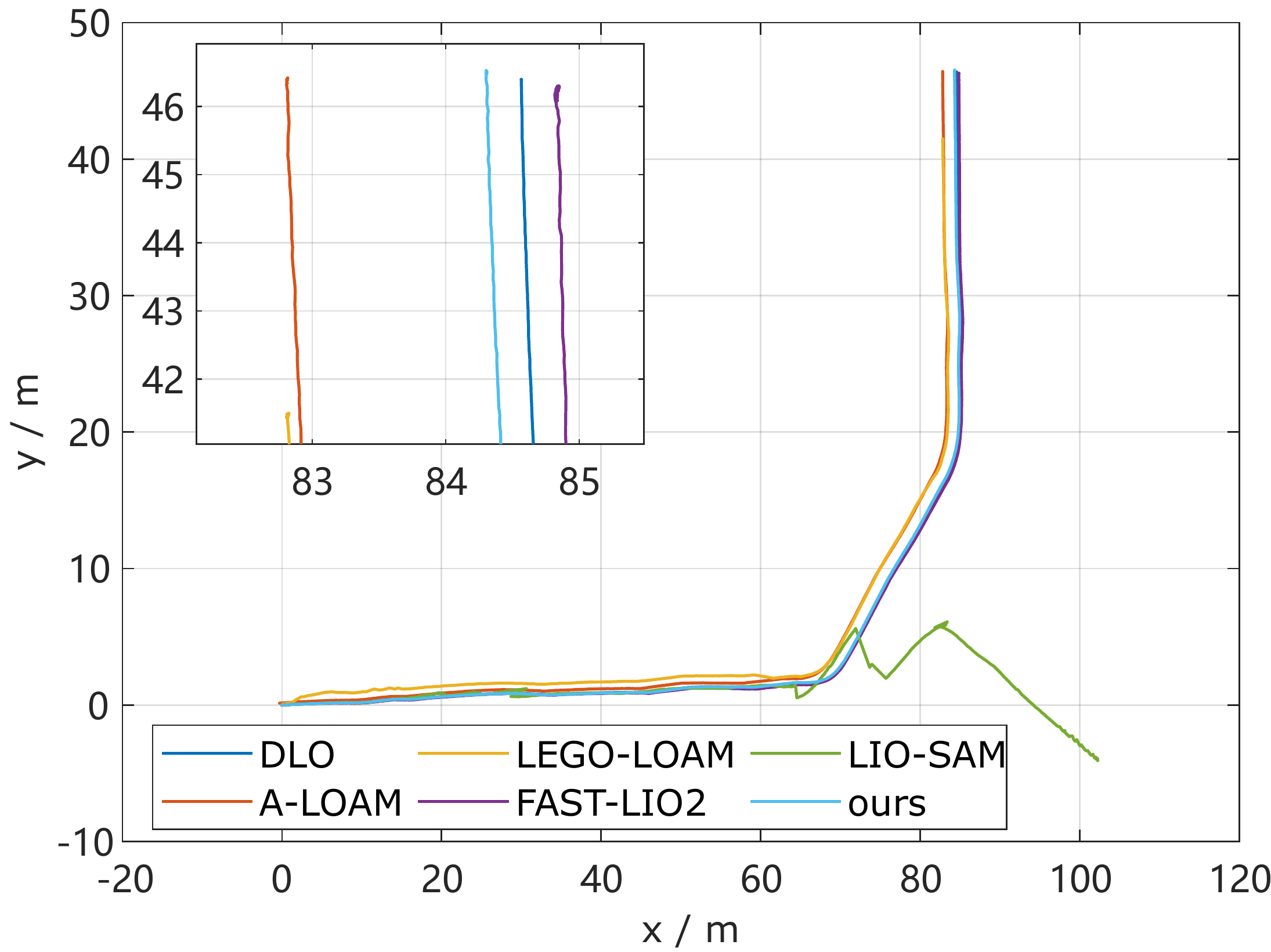}}\quad%1.8
    \subfloat[z-axis and orientation results]{\includegraphics[width=3.4in]{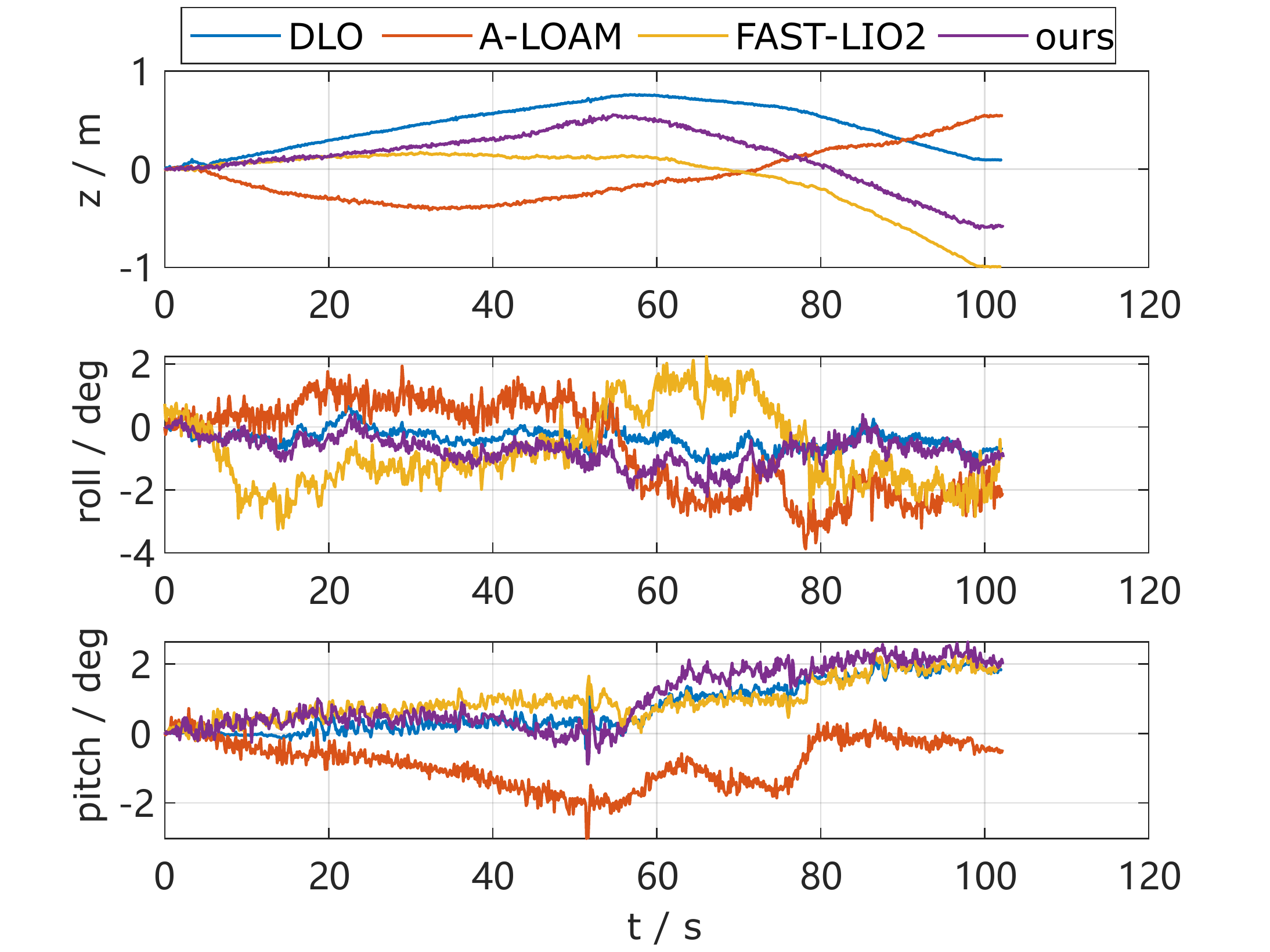}}%\vspace{12pt}
    \caption{Trajectory and orientation results from different methods in exp2.}
    \label{1xy}
\end{figure}

\begin{figure}
\centering
%1.1
\subfloat[ours]{\includegraphics[width=1.7in]{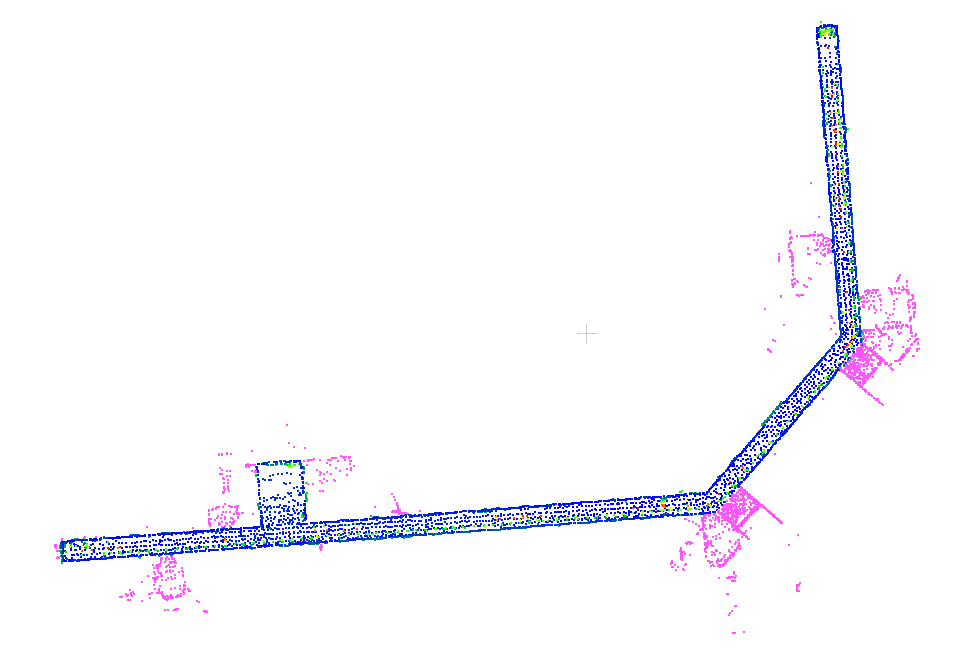}}
\subfloat[A-LOAM]{\includegraphics[width=1.7in]{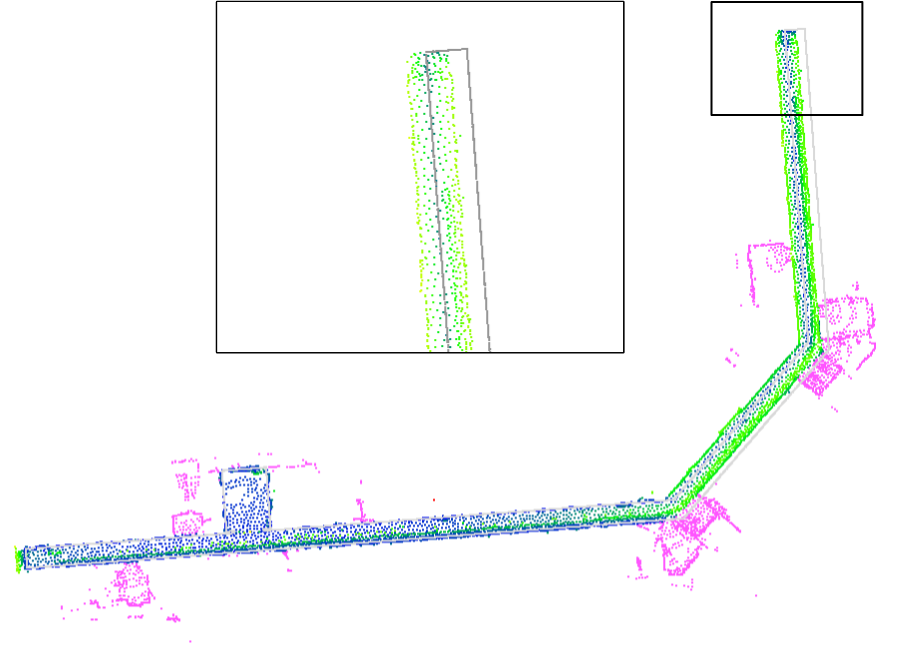}}\quad
\subfloat[DLO]{\includegraphics[width=1.7in]{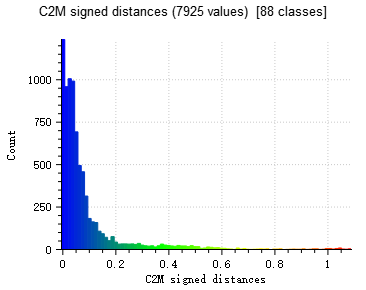}}
\subfloat[A-LOAM]{\includegraphics[width=1.7in]{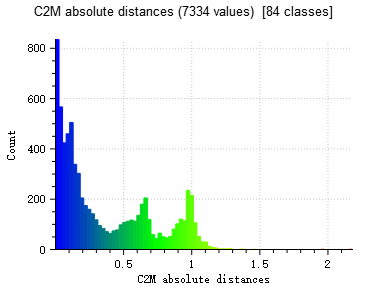}}\quad
\subfloat[FAST-LIO2]{\includegraphics[width=1.7in]{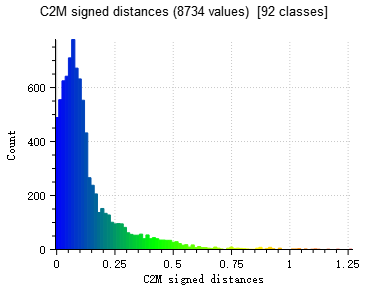}}
\subfloat[ours]{\includegraphics[width=1.7in]{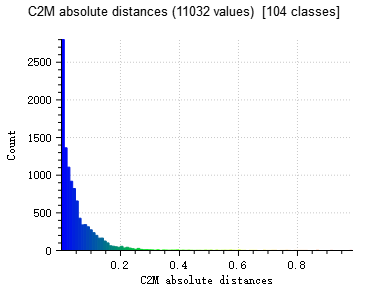}}
\caption{Mapping results in exp2. (a) is the mapping result of the RaLI-Multi. (b) is the mapping result of A-LOAM. The drifted area is detailed in partial enlargement and the gray points are sampled from the 3D model established in CATIA. (c)-(f) are C2M distances from DLO, A-LOAM, FAST-LIO2 and ours.}
\label{1map}
\end{figure}

\begin{table*}[!t]
\begin{center}
\caption{Mapping results of different methods in three experiments.}
\label{tmap}
\begin{tabular}{ c c c c c c c c c }
\hline
&  & DLO & A-LOAM & LeGO-LOAM & FAST-LIO2 & LIO-SAM & ours \\
\hline
\multirow{2}*{exp1} & mean distance & 0.0730781 & 0.277893 & \multirow{2}*{drifted} & \multirow{2}*{drifted} & \multirow{2}*{drifted} & $\textbf{0.0524321}$ \\
\cline{2-4} \cline{8-8}
~ & std distance & 0.143915 & 0.359658 &  &  & & $\textbf{0.121798}$ \\
\hline
\multirow{2}*{exp2} & mean distance & 0.0884124 & 0.366741 & \multirow{2}*{drifted} & 0.133056 & \multirow{2}*{drifted} & $\textbf{0.0548707}$ \\
\cline{2-4} \cline{6-6} \cline{8-9}
& std distance & 0.144602 & 0.345045 &  & 0.142586 &  & $\textbf{0.078375}$ \\
\hline
\multirow{2}*{exp3} & mean distance & 0.218776 & 0.189309 & \multirow{2}*{failed} & \multirow{2}*{failed} & \multirow{2}*{failed} &  $\textbf{0.0991533}$ \\
\cline{2-4} \cline{9-9}
& std distance & 0.205369 & 0.198182 &  &  &  & $\textbf{0.114884}$ \\
\hline
\end{tabular}
\end{center}
\end{table*}

Next, we evaluate different methods in a long corridor-like environment. As a result that the narrow environment constrains the number of vehicles, we evaluate the RaLI-Multi system with two vehicles and they play `the anchor role' in turn during the exploration. Since the environment in exp2 is longer and narrower than that in exp1, which is difficult to arrange our distributed localization system, we only evaluate the accuracy of the point cloud map.

Trajectories are demonstrated in Fig. \ref{1xy}. LIO-SAM degenerated and failed at the end of the first long corridor, around (65, 2) in Fig. \ref{1xy} (a). Although A-LOAM and LeGO-LOAM resisted degradation to some extent, they drifted in different axes, with A-LOAM mostly in the x-axis and LeGO-LOAM in XY-axes. Both of them drift at the same place as LIO-SAM. Moreover, A-LOAM drifts in the z-axis shortly after the beginning of this experiment. The rest three methods, DLO, FAST-LIO2, and the RaLI-Multi show similar results.

Then, we evaluate the mapping results. The results of our RaLI-Multi and A-LOAM are shown in Fig. \ref{1map} (a) and (b), respectively. In the zoom area of \ref{1map} (b), gray points are point clouds sampled from the 3D reference model and green points show significant offsets from the reference. Due to large drifts of A-LOAM, C2M distances also distribute widely. As we can see from Fig. \ref{1map} (d), A-LOAM shows two local peaks near 0.65m and 1m. Although FAST-LIO2 and DLO show similar performance in table \ref{tmap}, the peak of the C2M distance histogram of FAST-LIO2 in Fig. \ref{1map} (e) locates away from zero. Among these methods, the first histogram of C2M distances of the RaLI-Multi occupies the most percentage, over 20\%, and most C2M distances are within 0.2m. Details of C2M distances are also shown in table \ref{tmap}.

\subsection{Evaluation in a Complex Environment}

\begin{figure}
    \centering
    \subfloat[benchmarks trajcetory]{\includegraphics[width=1.5in]{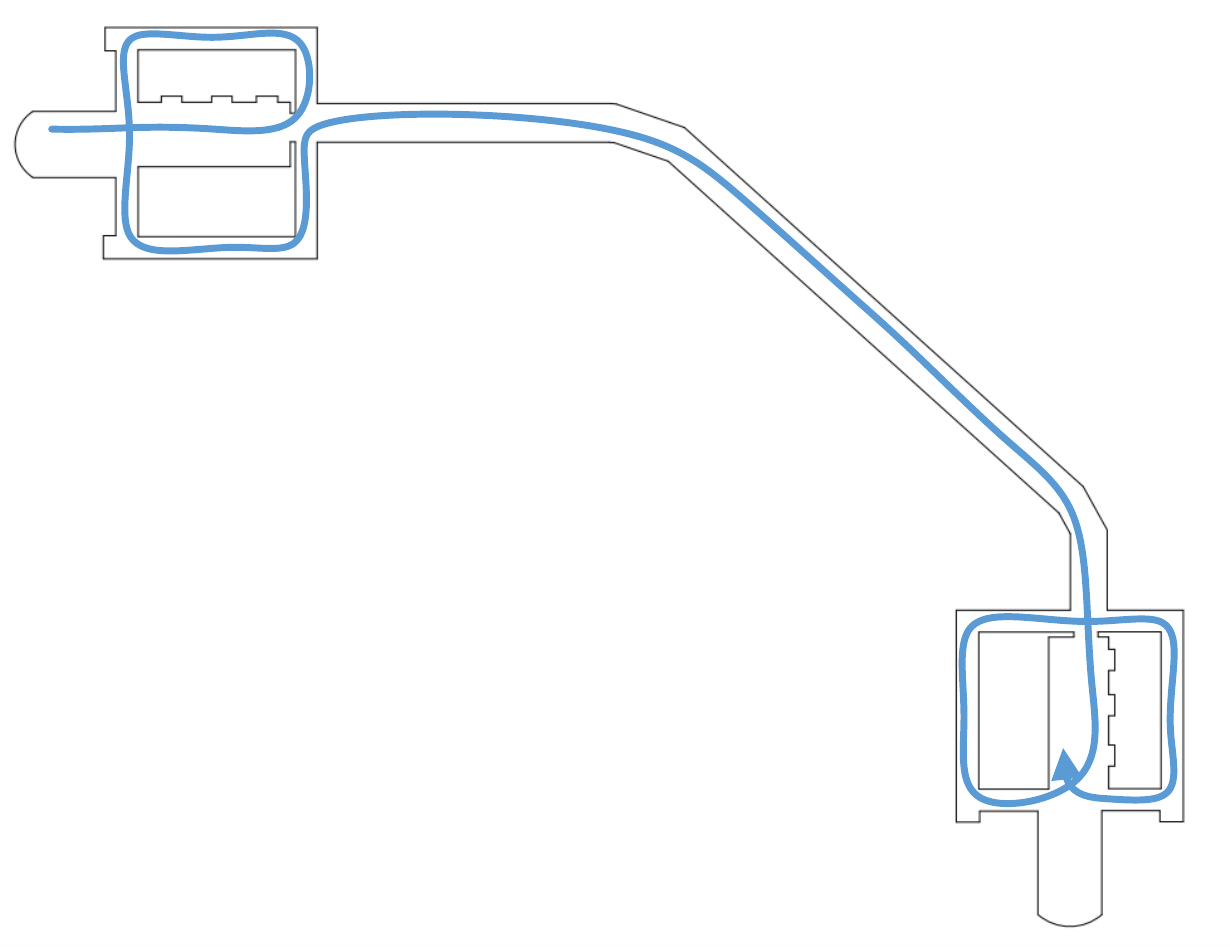}}
    \subfloat[round 1]{\includegraphics[width=1.5in]{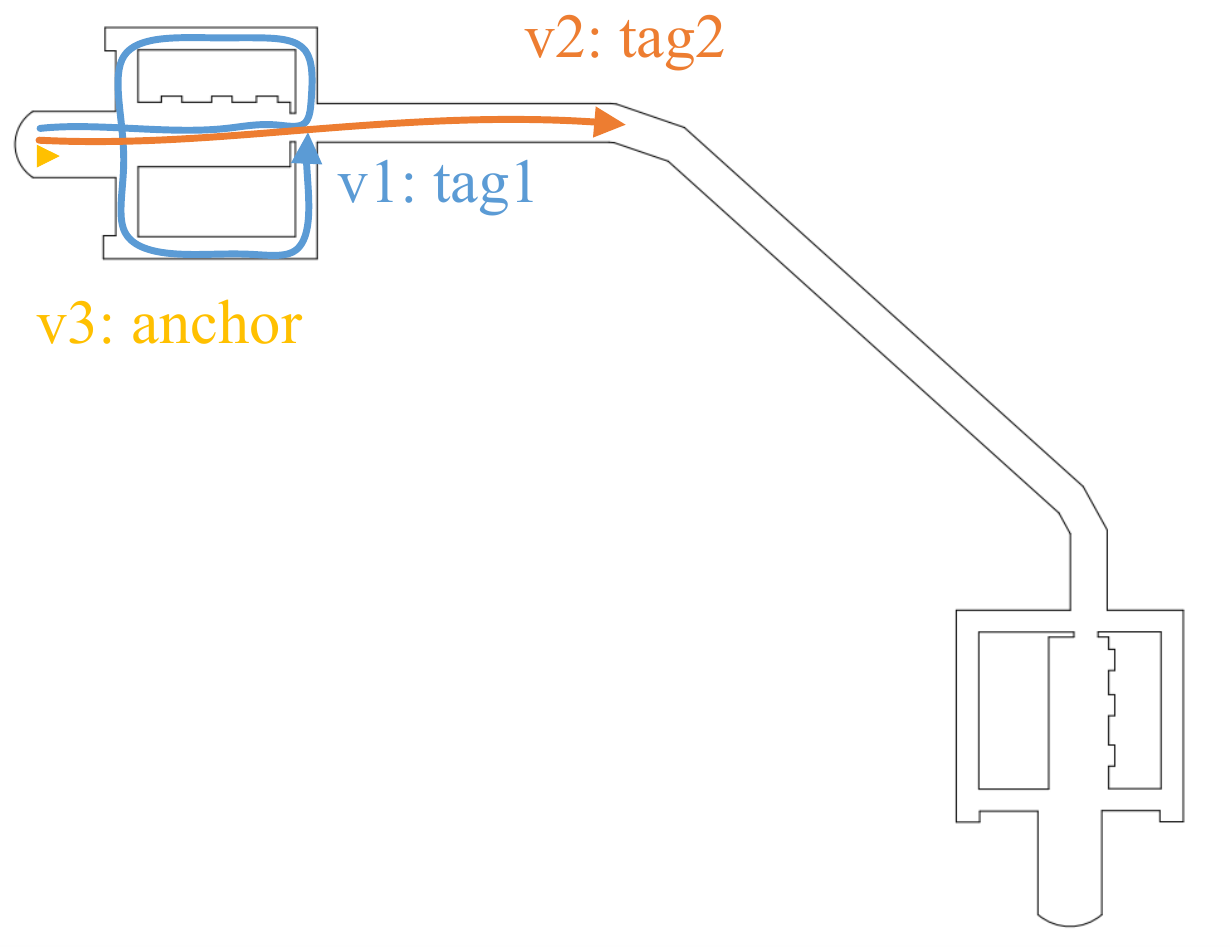}}\quad
    \subfloat[round 2]{\includegraphics[width=1.5in]{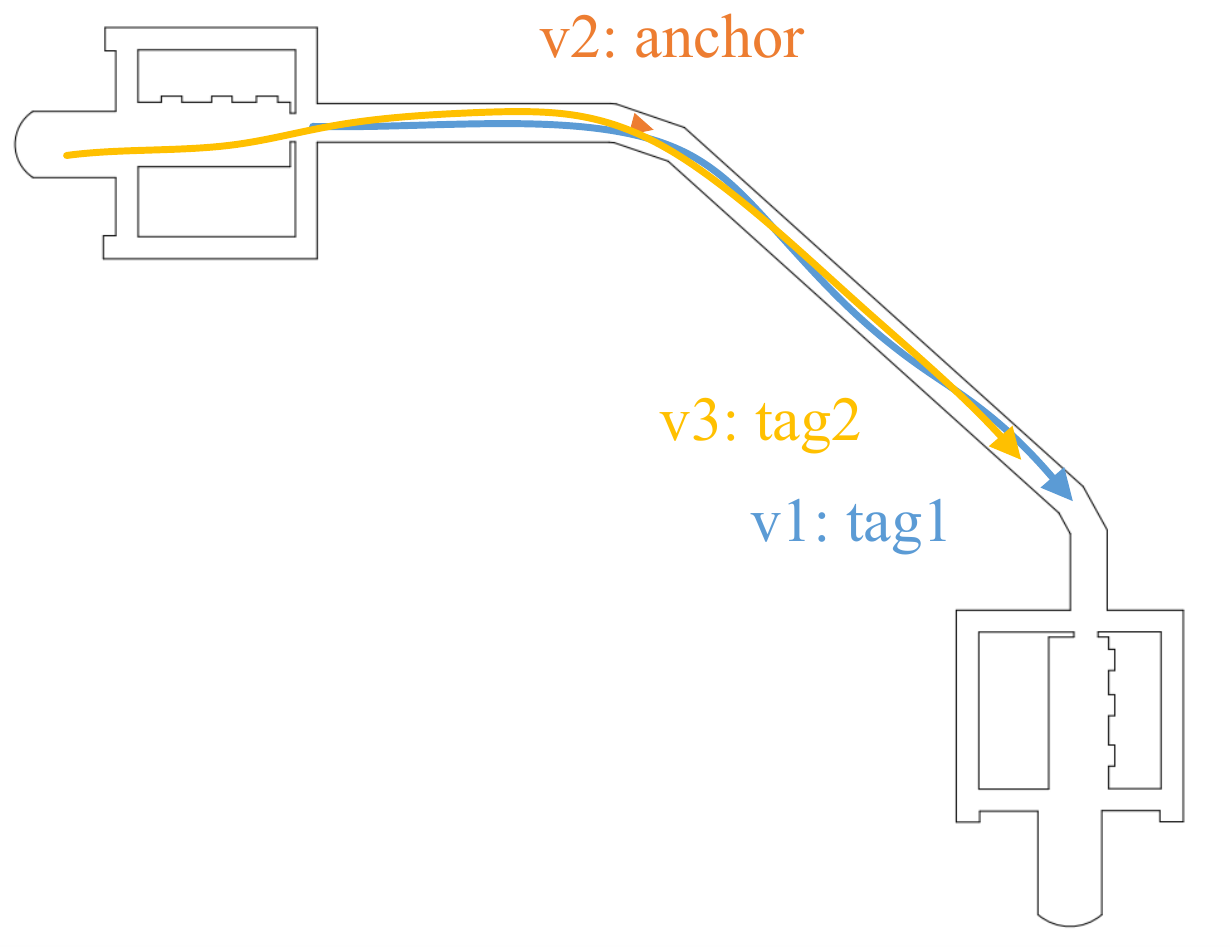}}
    \subfloat[round 3]{\includegraphics[width=1.5in]{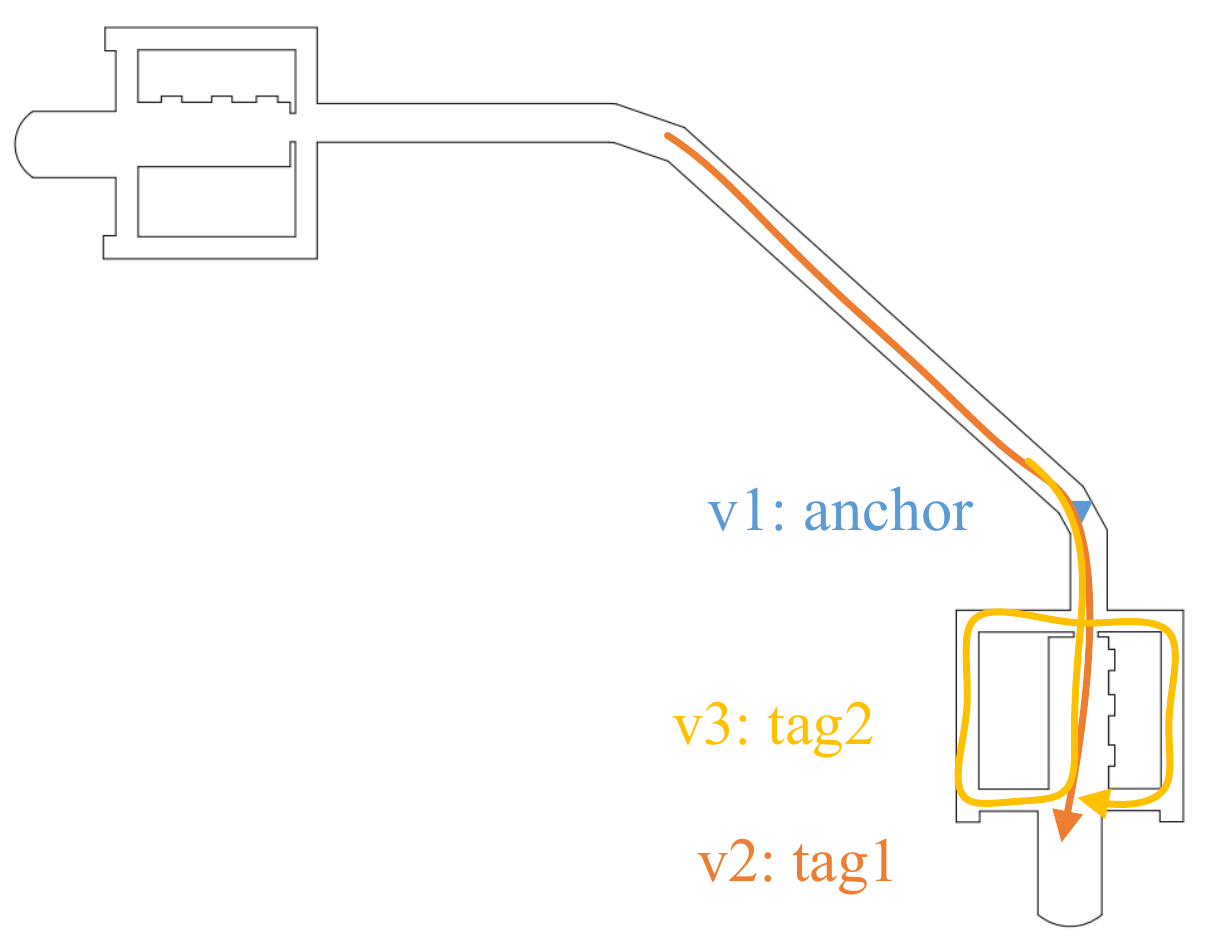}}
    \caption{Trajectory schematic of vehicles in the benchmarks and the RaLI-Multi. (a) The trajectory schematic of the vehicle in the benchmarks. (b)-(d) Trajectory schematics of vehicles in the RaLI-Multi in three exploration rounds and each color represents a vehicle.}
    \label{2route}
\end{figure}

\begin{figure}[!t]
    \centering
\begin{minipage}[b]{1.5in}
    \centering
    \subfloat[DLO]{\includegraphics[height=1.3in]{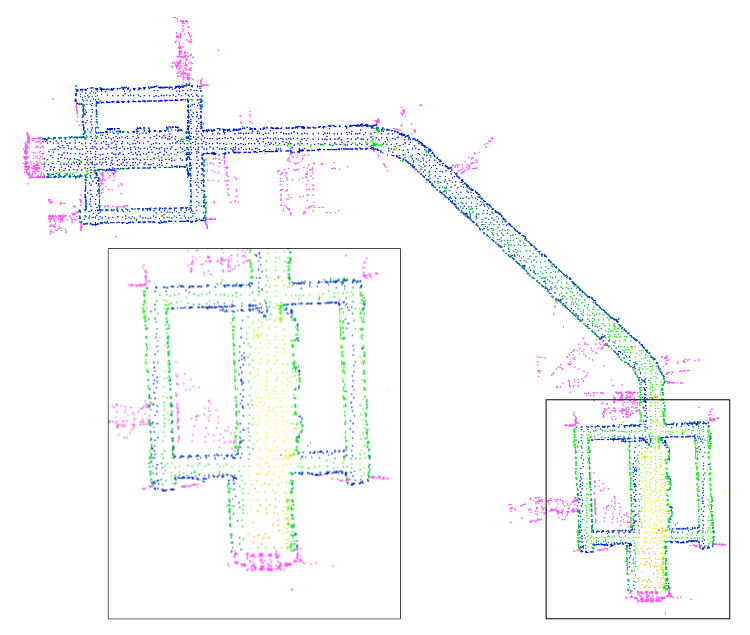}}\\
    \subfloat[A-LOAM]{\includegraphics[height=1.3in]{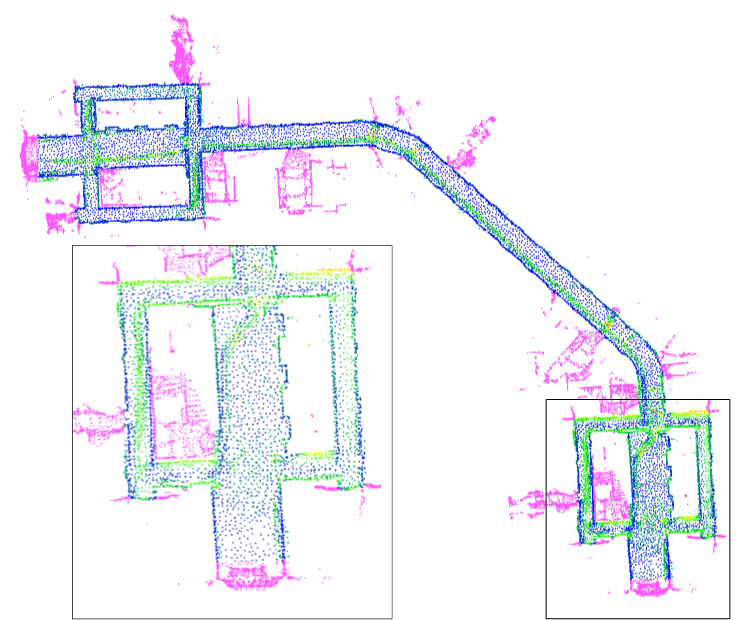}}\\
    \subfloat[ours]{\includegraphics[height=1.3in]{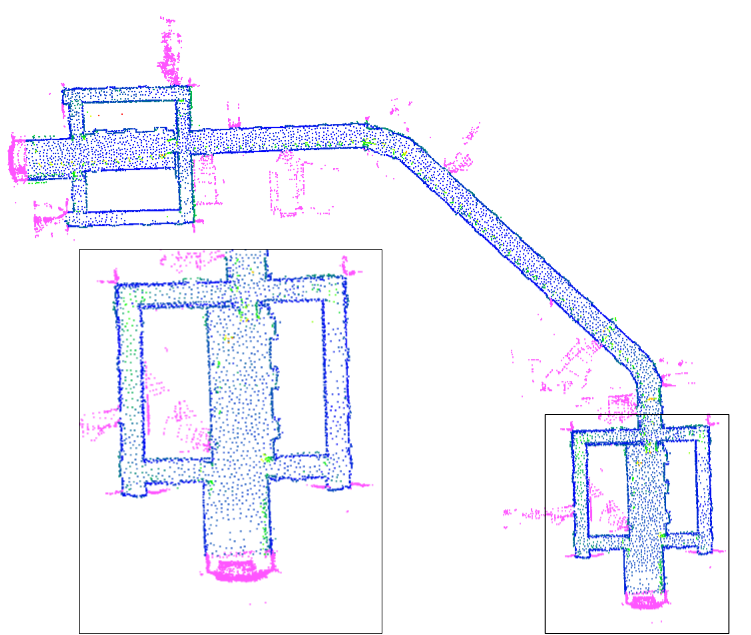}}
\end{minipage}
%\medskip
\begin{minipage}[b]{1.5in}
    \centering
    \subfloat[DLO]{\includegraphics[height=1.3in]{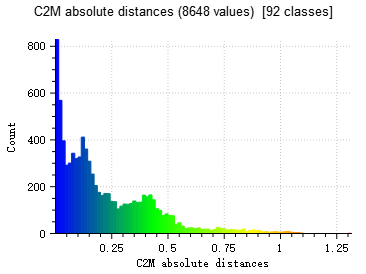}}\\
    \subfloat[A-LOAM]{\includegraphics[height=1.3in]{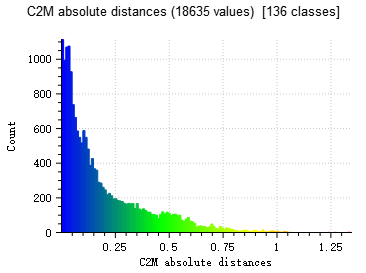}}\\
    \subfloat[ours]{\includegraphics[height=1.3in]{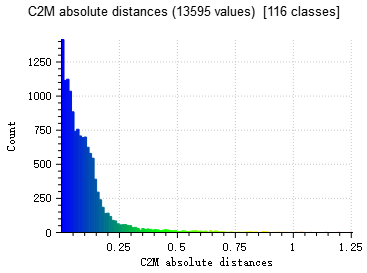}}
\end{minipage}

%    \subfloat[DLO]{\includegraphics[width=1.1in]{img/exp2/mdlo2.png}}
%    \subfloat[A-LOAM]{\includegraphics[width=1.1in]{img/exp2/maloam2.png}}
%    \subfloat[ours]{\includegraphics[width=1.1in]{img/exp2/mlio2.png}}\quad
%    \subfloat[DLO]{\includegraphics[width=1.1in]{img/exp2/dlo.png}}
%    \subfloat[A-LOAM]{\includegraphics[width=1.1in]{img/exp2/aloam.png}}
%    \subfloat[ours]{\includegraphics[width=1.1in]{img/exp2/lio.png}}
    \caption{Mapping results in exp3. (a)-(c) are mapping results from DLO, A-LOAM and ours. (d)-(f) are corresponding C2M distances.}
    \label{2map}
\end{figure}

Finally, we evaluate different methods in a more complex environment. The proposed RaLI-Multi consisting of three vehicles and the benchmarks, DLO, A-LOAM, LeGO-LOAM, FAST-LIO2, and LIO-SAM, apply different trajectories, as shown in Fig. \ref{2route}. %

Fig. \ref{2route} (b)-(d) show three exploration rounds of the RaLI-Multi in this experiment. In the first round, vehicles 1 and 2 are two tag vehicles and vehicle 3 is the initial anchor vehicle. Vehicle 1 explores the initial place while vehicle 2 explores in the right direction. The anchor, vehicle 3, receives information from vehicles 1 and 2 to perform the initialization and global optimization. After two tag vehicles finish their exploration, vehicle 2 is selected as the next anchor because only vehicle 2 has frontiers among tag vehicles. In the second round, both tag vehicles are heading in the bottom right direction. At the corner of the corridor, vehicle 1 stops and is selected as the next anchor. In the third round, vehicle 3 explores the bottom right area while vehicle 2 heads toward the end of the corridor.

The mapping results are shown in table \ref{tmap} and Fig. \ref{2map}, where LeGO-LOAM, FAST-LIO2, and LIO-SAM failed in different places. We then evaluate the mapping results of the rest three methods. As shown in Fig. \ref{2map} (a)-(c), DLO and A-LOAM have higher C2M distances at the labeled area than the RaLI-Multi. From the zoom area of point cloud maps, A-LOAM drifts more in horizontal directions (XY axes) than DLO, demonstrated as green points, while DLO shows more errors in the vertical direction (Z axis), represented in green and yellow points. Combining Fig. \ref{2map} (d)-(f), most C2M distance histograms of the RaLI-Multi are within 0.2m while DLO and A-LOAM still have local peaks around 0.4m and 0.5m, respectively. Statistics of mapping results shown in table \ref{tmap} also demonstrate the proposed method achieves a better result than the state-of-the-art.

\section{CONCLUSION}
\label{sec c}

In this paper, we propose a range-aided LiDAR-inertial multi-vehicle mapping system for a large-scale environment with degeneration. The multi-metric weights LiDAR-inertial front-end assigns weights to each feature point, based on the distance and the neighbors of them, and the kinematics of the vehicle, which improves the performance in narrow and degenerate environments. The degeneration detection module can online monitor the degeneration via the distribution of normal vectors of feature points. The degenerate correction module can compensate for the LiDAR-inertial odometry along the degenerate direction.
The dynamically centralized multi-vehicle system can robustly and flexibly operate in various complex and degenerate environments.

Three experiments demonstrate that: 1) the proposed LiDAR-inertial front-end can resist degeneration and achieve better mapping results; 2) with the help of degeneration detection and correction, the proposed multi-vehicle system can obtain a low-drift global map in degenerate environments; 3) compared with the state-of-the-art, the RaLI-Multi is more robust in the three experiments.

%\begin{thebibliography}{unsrt}
\bibliographystyle{IEEEtran}

\bibliography{ref}

%\end{thebibliography}

%\newpage

% {Biography Section}
% your own custom macro containing the $\backslash${\tt{includegraphics}} command to make things
%\bf{If you include a photo:}\vspace{-33pt}
%\begin{IEEEbiography}[{\includegraphics[width=1in,height=1.25in,clip,keepaspectratio]{fig1}}]{Michael Shell}
%Use $\backslash${\tt{begin\{IEEEbiography\}}} and then for the 1st argument use $\backslash${\tt{includegraphics}} to declare and link the author photo.
%Use the author name as the 3rd argument followed by the biography text.
%\end{IEEEbiography}

\vfill

\end{document}